\documentclass[11pt]{article}
\usepackage{algorithm}  
\usepackage{algpseudocode}
\usepackage[margin=1in]{geometry}
\usepackage[numbers,sort&compress]{natbib}
\usepackage{bm}

\usepackage[utf8]{inputenc} 
\usepackage[T1]{fontenc}    
\usepackage{hyperref}       
\usepackage{url}            
\usepackage{booktabs}       
\usepackage{amsfonts}  

\usepackage{nicefrac}       
\usepackage{microtype}      
\usepackage{xcolor}         
\usepackage{enumitem}
\usepackage{multirow}
\usepackage{mathtools}
\usepackage{amssymb}
\usepackage{colortbl}
\usepackage{graphicx}
\usepackage{url}
\usepackage{pgfplots}  
\usepackage{caption}      
\usepackage{subcaption} 
\usepackage{amsmath}
\usepackage{booktabs}
\usepackage{wrapfig}
\usepackage{float}
\usepackage{enumitem}

\newtheorem{theorem}{Theorem}[section]

\newtheorem{lemma}[theorem]{Lemma}

\newtheorem{definition}[theorem]{Definition}
\newtheorem{assumption}[theorem]{Assumption}
\newtheorem{remark}[theorem]{Remark}
\makeatletter
\renewcommand*{\@fnsymbol}[1]{\ensuremath{\ifcase#1\or \dagger\or \ddagger
\else\@ctrerr\fi}}
\makeatother

\title{Decentralized Low-Rank Fine-Tuning of Large Language Models}
\author{
        \textbf{Sajjad Ghiasvand}$^1$ \ \ 
        \textbf{Mahnoosh Alizadeh}$^1$ \ \  
        \textbf{Ramtin Pedarsani}$^1$
       \\ 
  Electrical and Computer Engineering Department, UC Santa Barbara$^1$ 
  \\
  {\tt \{sajjad,alizadeh,ramtin\}@ucsb.edu} 
}
\date{}
\begin{document}

\maketitle
\begin{abstract}
\normalsize

While parameter-efficient fine-tuning (PEFT) techniques like Low-Rank Adaptation (LoRA) offer computationally efficient adaptations of Large Language Models (LLMs), their practical deployment often assumes centralized data and training environments. However, real-world scenarios frequently involve distributed, privacy-sensitive datasets that require decentralized solutions. Federated learning (FL) addresses data privacy by coordinating model updates across clients without sharing raw data. While most federated fine-tuning methods adopt \textit{centralized FL}, which relies on a parameter server for aggregating model updates—introducing potential bottlenecks and communication constraints—\textit{decentralized FL} enables direct peer-to-peer communication among clients, bypassing the need for a server as an intermediary. Despite its advantages, decentralized fine-tuning for LLMs remains largely unexplored in the literature. To address this gap, we introduce \texttt{Dec-LoRA}, a decentralized fine-tuning algorithm based on LoRA.
We conduct extensive experiments using BERT and LLaMA-2 models to benchmark \texttt{Dec-LoRA} against centralized LoRA and several other popular PEFT approaches in decentralized settings. Our results demonstrate that \texttt{Dec-LoRA} consistently achieves performance on par with centralized LoRA under various conditions, including data heterogeneity and quantization constraints. Furthermore, we provide a rigorous theoretical analysis, proving that under standard assumptions of smoothness and bounded gradients for non-convex functions, \texttt{Dec-LoRA} converges to a stationary point at a rate of $ O\left( \frac{1}{T^{1/2}} \right) $. These findings highlight the potential of \texttt{Dec-LoRA} for scalable LLM fine-tuning in decentralized environments.


\end{abstract}

\section{Introduction}

The advent of Large Language Models (LLMs) such as GPT-4~\cite{achiam2023gpt}, LLaMA~\cite{touvron2023llama}, and BERT~\cite{devlin2018bert} has revolutionized artificial intelligence by enabling remarkable capabilities in tasks such as translation and summarization~\cite{bommasani2021opportunities}, powered by sophisticated architectures like Transformers~\cite{vaswani2017attention}. These versatile models can be fine-tuned for domain-specific applications using targeted datasets~\cite{howard2018universal}, showcasing their adaptability across diverse fields. However, the sheer scale of these models, often comprising billions of parameters, makes complete fine-tuning computationally prohibitive and prone to overfitting. To address this, parameter-efficient fine-tuning (PEFT) techniques—such as Adapters~\cite{houlsby2019parameter}, Prompt-Tuning~\cite{lester2021power},  LoRA~\cite{hu2021lora}—have emerged as practical solutions. These approaches selectively adjust only a fraction of the model parameters while keeping the rest static, significantly cutting computational demands without compromising performance~\cite{ding2023parameter}. Among these, LoRA is preferred in certain applications and has been shown to have excellent efficiency, making it the focal point of our study.

Traditional PEFT methods often assume that LLMs are fine-tuned using data from a single machine or client. However, in real-world scenarios, sensitive data sets, such as medical records or legal documents, are frequently distributed across multiple devices~\cite{manoel2023federated,shoham2023federated}. Privacy concerns make centralizing such data impractical, creating the urgent need for fine-tuning techniques capable of adapting LLMs at the edge while maintaining strict data privacy. In response to this challenge, Federated Learning (FL)~\cite{mcmahan2017communication} emerges as a powerful solution by ensuring sensitive information remains on local devices throughout the training process. Instead of transferring raw data to a centralized server for training, FL enables clients to update model parameters locally and share only aggregated information, such as gradients or parameters~\cite{mcmahan2017communication}. Consequently, FL has been seamlessly integrated into PEFT approaches~\cite{zhang2023fedpetuning,fan2023fate,zhao2023fedprompt,ghiasvand2024communicationllm}, with federated fine-tuning of LoRA receiving particular attention for its ability to efficiently balance privacy, communication overhead, and model adaptability across different clients~\cite{babakniya2023slora,yan2024federa,cho2023heterogeneous,bai2024federated,wang2024flora,kuo2024federated,sun2024improving,chenrobust}. 

Almost all previous work on federated fine-tuning focuses on \textit{centralized FL}, which relies on a centralized server to coordinate the aggregation of model updates. This dependency poses challenges, particularly in scenarios where communication resources are limited or where a centralized server introduces potential bottlenecks. Another FL architecture, called \textit{decentralized FL}, enables direct peer-to-peer communication among clients, bypassing the need for a server as an intermediary~\cite{yuan2024decentralized}, while still preserving the key advantages of centralized FL. Recent advances have demonstrated the effectiveness of decentralization in LLM-based multi-agent systems, facilitating scalable and robust collaboration among distributed agents~\cite{guo2024large,chen2024scalable}. Despite its broader applicability and critical role in emerging applications, decentralized fine-tuning for LLMs remains largely unexplored in the literature. In this work, we address this gap by proposing a decentralized fine-tuning algorithm and provide both empirical evidence and theoretical guarantees of its effectiveness.


Before delving into details, we summarize our contributions: \textbf{(1)} We introduce \texttt{Dec-LoRA}, which, to the best of our knowledge, is the first algorithm designed to fine-tune LLMs in a decentralized setting. \textbf{(2)} We propose a tractable optimization metric and prove that, under standard assumptions such as smoothness and bounded gradients of the local functions, \texttt{Dec-LoRA} converges to a stationary point with respect to this metric at a rate of $O(\frac{1}{T^{1/2}})$.
\textbf{(3)} We benchmark \texttt{Dec-LoRA} against several popular PEFT approaches in decentralized settings and show that it consistently achieves superior accuracy and faster convergence on average across various tasks and settings. \textbf{(4)} We conduct extensive experiments using BERT and LLaMA-2 family models, comparing centralized LoRA and \texttt{Dec-LoRA} under diverse settings, including data heterogeneity and quantization constraints. The results show that \texttt{Dec-LoRA} is an effective and practical solution for decentralized fine-tuning of LLMs.


\section{Preliminaries}
\subsection{Low-Rank Adaptation}
Low-Rank Adaptation (LoRA)~\cite{hu2021lora} is one of the most promising PEFT methods, enabling effective fine-tuning of LLMs by freezing the entire model and adding low-rank trainable matrices in each layer. LoRA has been shown to outperform other PEFT methods, even in federated learning settings~\cite{kuang2024federatedscope}.

In LoRA, for a pre-trained weight matrix $ W_0 \in \mathbb{R}^{d_1\times d_2} $, the weight update is performed by a low-rank decomposition:
\begin{equation}\label{eq.1}
W_0+\Delta W=W_0+B A,
\end{equation}
where the training occurs on matrices $ A \in \mathbb{R}^{r\times d_2} $ and $ B \in \mathbb{R}^{d_1\times r} $, with $r \ll \min (d_1, d_2)$. Throughout the paper, we refer to $r$ as the \textit{rank} of LoRA, which is typically selected from $\{2,4,8,16\}$.

Beyond good performance, the low number of trainable parameters makes LoRA a practical solution for decentralized fine-tuning of language models, where clients have limited training resources and communication between clients is costly.

\subsection{Decentralized Fine-Tuning via LoRA}
We consider a connected network of $ n $ clients, denoted by $ \mathcal{C} = \{c_1, \dots, c_n\} $, with edges $ \mathcal{E} \subseteq \mathcal{C} \times \mathcal{C} $ representing the communication links between clients. The network collaboratively aims to solve the following optimization problem:
\begin{align}
\min_{A,B} \left[ f(W_0 +BA) := \frac{1}{n} \sum_{i=1}^n f_i(W_0+ BA)\right],\label{lora_equation}
\end{align}
where $ W_0 $ is the pre-trained model that is shared and fixed across all clients,
and the local functions $f_i:\mathbb{R}^{d_1\times d_2} \rightarrow \mathbb{R}$ are distributed among $n$ clients and are given in stochastic form:
\begin{align}
    f_i(W_0+BA) = \mathbb{E}_{\xi_i \sim \mathcal{D}_i}[F_i(W_0 + BA; \xi_i)].
\end{align}
Here, the expectation is taken with respect to randomly selected sample set $ \xi_i \sim \mathcal{D}_i $, where $ \mathcal{D}_i $ denotes the local data distribution specific to client $ c_i $. Standard empirical risk minimization is an important special case that is covered in this setting.

In this decentralized setting, the clients communicate with each other along the edges $ e \in \mathcal{E} $, which means that each client can communicate with its neighboring clients. Furthermore, each edge in the graph is associated with a positive mixing weight, and we denote the mixing matrix by $ Q=\left[q_{i j}\right] \in \mathbb{R}^{n \times n} $.
\subsection{Notation}
$\|A\|_F$ denotes Frobenius norm of $A$. We also define the following quantities:
\begin{align} \bar{A} := \frac{1}{n}\sum_{i=1}^n A_i, \quad \bar{B} := \frac{1}{n}\sum_{i=1}^n B_i, \quad W:= W_0 + BA. \nonumber \end{align}
With a slight abuse of notation, we denote \( f(W) \) as \( f(BA) \) and use \( f(W) \) and \( f(BA) \) interchangeably. Additionally, we define:
$\Tilde{\nabla}f_i(W) := \nabla F_i(W;\xi_i).$

\section{Proposed Algorithm}



\begin{figure}[t]
    \centering 
    \scriptsize
    \begin{subfigure}[b]{0.46\textwidth} 
        \centering
        \includegraphics[width=\textwidth]{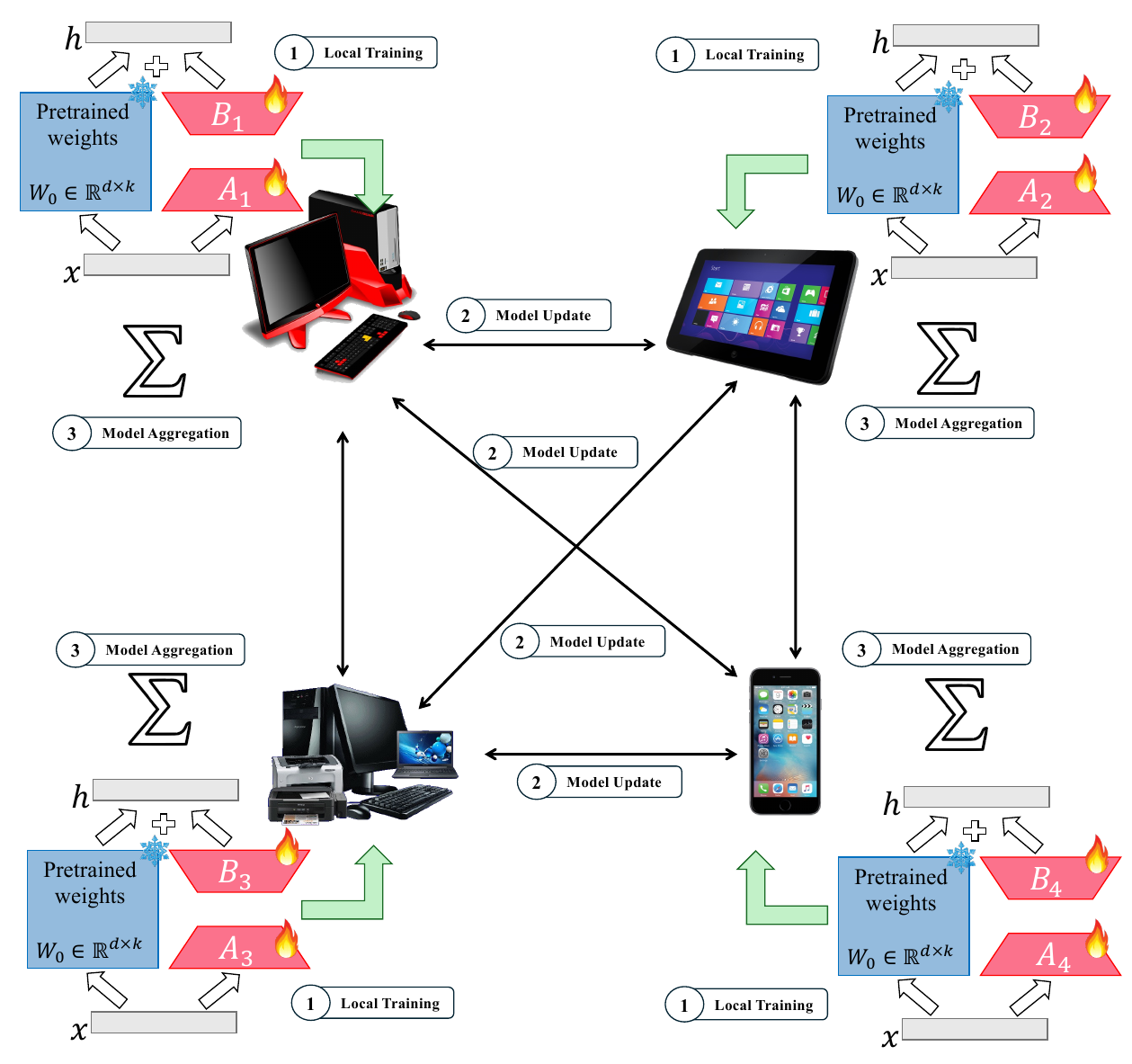}
        \caption{\includegraphics[height=1em]{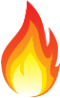}: Trainable Parameters, \includegraphics[height=1em]{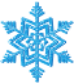}: Frozen Parameters}
        \label{fig:method1}
    \end{subfigure}
    \hspace{0.25cm}
    \begin{subfigure}[b]{0.50\textwidth}
        \centering
        \includegraphics[width=\textwidth]{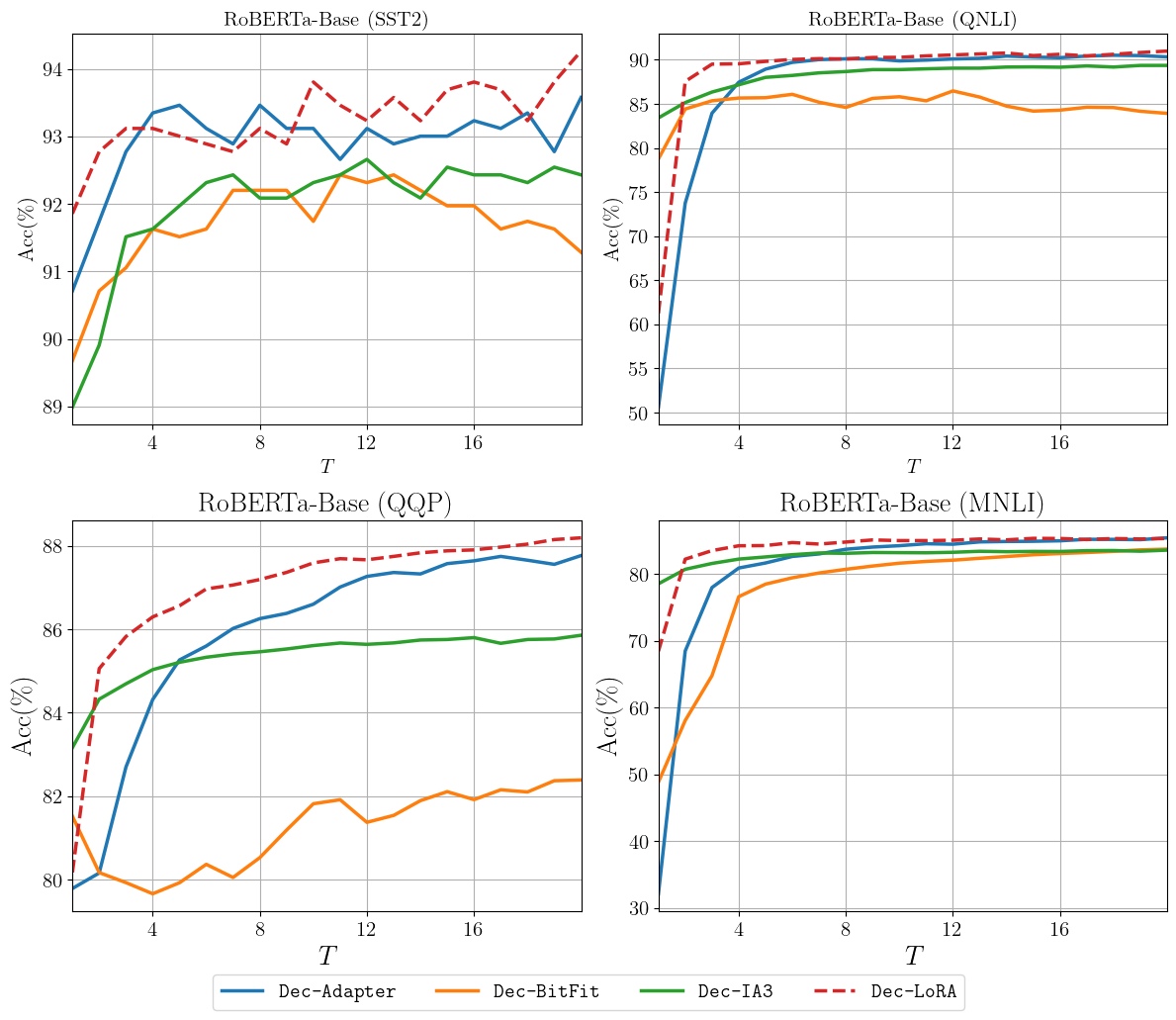}
        \caption{}
        \label{fig:method2}
    \end{subfigure}
    
    \caption{(a): Illustration of the \texttt{Dec-LoRA} algorithm. The process includes three stages: (1) local training of low-rank matrices $A$ and $B$ on each client for $K$ iterations using their private data, (2) communication of updated parameters between neighboring clients in the network, and (3) aggregation of received updates by each client using the mixing matrix $Q$ to compute the next round's parameters. (b) Convergence speed of decentralized PEFT methods using the Ring topology.
  }
  \label{fig:method}
\end{figure}

       
In this section, we introduce our proposed \texttt{Dec-LoRA} algorithm.
 At the start of the fine-tuning process, the full model's architecture and initial weights ($A^{(0)} \sim \mathcal{N}(0, \sigma^2)$, $B^{(0)}=\boldsymbol{0}$) are distributed to all clients in the set $\mathcal{C} = \{c_1, \cdots, c_n\}$. \texttt{Dec-LoRA} operates across $T$ communication rounds, where each client performs $K$ local updates on its trainable LoRA parameters in each round.

During a communication round \( t \), each client \( c_i \in \mathcal{C} \) initializes its local LoRA matrices with the obtained LoRA matrices from the previous round, \( A_i^{(t)} \) and \( B_i^{(t)} \), and then performs local training on its local dataset for \( K \) local updates: 
\begin{gather}
    A_i^{(t)+k+1} = A_i^{(t)+k} - \eta \Tilde{\nabla}_A f_i(W_i^{(t)+k}),  \quad
    B_i^{(t)+k+1} = B_i^{(t)+k} - \eta \Tilde{\nabla}_B f_i(W_i^{(t)+k}), \label{eq:updata}
\end{gather}
where \( \eta \) is the learning rate, and \( W_i^{(t)+k} \), \( A_i^{(t)+k} \), \( B_i^{(t)+k} \) refer to the weight matrix and LoRA matrices for client \( c_i \) during communication round \( t \) and local update \( k \).

Once the updates are complete, each client transmits its updated parameters, \( A_i^{(t)+K} \) and \( B_i^{(t)+K} \), to its neighboring clients. The clients then aggregate the parameters received from their neighbors using the mixing matrix \( Q \). The updated parameters for client \( c_i \) are computed as:
\begin{align}
    A_i^{(t+1)} = \sum_j q_{ij} A_j^{(t)+K},\quad
    B_i^{(t+1)} = \sum_j q_{ij} B_j^{(t)+K}. \label{eq:average}
\end{align}
 The steps of the \texttt{Dec-LoRA} algorithm are illustrated in Fig.~\ref{fig:method1}. Moreover, \texttt{Dec-LoRA} pipeline can be found in Appendix \ref{pipeline}.

\section{Convergence Analysis}
    First, we briefly explain the key theoretical challenges and contributions of this work. For centralized LoRA, strong theoretical guarantees have yet to be discovered, but several important directions have been explored (see Appendix \ref{theory_lora}). In federated learning, providing theoretical guarantees is even more challenging since \( A \)-updates are aggregated separately from \( B \)-updates, whereas the "mathematically correct" update would be to average the products \( BA \) \cite{sun2024improving}. What we expect, under the assumption of non-convexity of the objective functions in federated learning, is to find a stationary point of the global objective function, i.e., showing that the following metric converges to zero as $T \rightarrow \infty$ \cite{fallah2020personalized,wang2020tackling}:
\begin{align}
    \frac{1}{T}\sum_{t=0}^{T-1}\mathbb{E}\left\| \frac{1}{n}\sum_{i=1}^n \nabla_W f_i\left(\frac{1}{n }\sum_{i=1}^n B_i^{(t)}A_i^{(t)}\right)\right\|_F^2,\label{federated_bound::1}
\end{align}
where \( W = W_0 + BA \). As mentioned earlier, deriving upper bounds for~\eqref{federated_bound::1} is challenging due to the separate updating and aggregation of \( A \) and \( B \). To address this, we introduce a more tractable and practically meaningful metric for federated LoRA-based algorithms, motivated by the structure of typical LoRA implementations, defined as follows:
{\small
\begin{align}
    \frac{1}{T}\sum_{t=0}^{T-1} &\mathbb{E}\left\| \frac{1}{n}\sum_{i=1}^n \nabla_{(A,B)} f_i\left(\frac{1}{n}\sum_{i=1}^n B_i^{(t)}\frac{1}{n}\sum_{i=1}^n A_i^{(t)}\right)\right\|_F^2.
\end{align}
}

This metric offers two key advantages: (1) it directly captures the stationarity condition with respect to the optimization variables \( (A, B) \), which represent the true degrees of freedom in low-rank adaptation; and (2) since the updates to \( A \) and \( B \) are aggregated separately, the search for a stationary solution is conducted over points of the form \( \bar{B}^{(t)}\bar{A}^{(t)} \), rather than the average product \( \frac{1}{n} \sum_{i=1}^n B_i^{(t)} A_i^{(t)} \).

Through a detailed convergence analysis, we prove that our decentralized LoRA algorithm converges to a stationary point with respect to this metric. The complete proofs are provided in Appendix~\ref{convergence}. Let us begin with a few assumptions.
\begin{assumption}\label{ass:smoothness}
    We assume that each local objective function $ f_i $ is $L$-smooth, that is, for all $W,W^{\prime} \in \mathbb{R}^{d_1\times d_2}$, we have
    \begin{align}
        \left\|\nabla f_i(W)-\nabla f_i(W^{\prime})\right\|_F \leq L\left\|W-W^{\prime}\right\|_F .
    \end{align}
\end{assumption}

\begin{assumption}\label{assm: bounded gradient}
We assume that the stochastic gradients are unbiased and that their expected squared norm remains uniformly bounded:
\begin{gather}
    \mathbb{E}_{\xi_i} \hspace{-0.1cm}\left[\nabla F_i(W; \xi_i)\right ] = \nabla f_i(W),\quad
    \mathbb{E}_{\xi_i}\hspace{-0.1cm}\left[\left\|\nabla F_i(W; \xi_i)\right\|^2_F\right] \leq G^2,
\end{gather}
for all $W \in \mathbb{R}^{d_1\times d_2}$ and $i=1, \ldots, n$, where $\xi_i$ represents a randomly sampled subset of training data from $i$-th client and $\mathbb{E}_{\xi_i}[\cdot]$  denotes the expectation over $\xi_i \sim \mathcal{D}_i$.
\end{assumption}

\begin{assumption}\label{assm: bounded matrices}
 Let $W_i^{(t)+k}=W_0+B_i^{(t)+k} A_i^{(t)+k}$ represent the model parameters for the $i$-th client during the $t$-th communication round and $k$-th local update. There exist constants $C_A>0$ and $C_B>0$ such that:
$
\|A_i^{(t)+k}\|_F  \leq C_A,
\|B_i^{(t)+k}\|_F  \leq C_B,\nonumber
$
for all $i=1, \cdots, n$, $t=0, \cdots, T-1$, and $k=0, \cdots, K-1$.
\end{assumption}
\begin{remark}
    For Assumption \ref{assm: bounded matrices}, based on the definition of the Frobenius norm,   
$
\|A\|_{F} = \sqrt{\sum_{i=1}^{m} \sum_{j=1}^{n} \left|a_{i j}\right|^2},  
$  
we observe that the stated inequalities hold as long as all elements in $A_i^{(t)+k}$ and $B_i^{(t)+k}$ remain finite. Note that the upper bounds $C_A$ and $C_B$ are independent of $t$ and $k$. However, they may depend on dimensions of $A_i^{(t)+k}$ and $B_i^{(t)+k}$, which are $r \times k$ and $d \times r$, respectively.
\end{remark}


As previously discussed, in our decentralized framework, clients communicate exclusively along the edges of a fixed communication graph that connects $n$ nodes. Each edge in this graph is associated with a positive mixing weight. These weights are collectively represented by the mixing matrix $ Q \in \mathbb{R}^{n\times n} $, which defines the network \textit{topology}.
\begin{assumption}\label{assm: mixing matrix}
    We assume that the mixing matrix $Q=\left[q_{i j}\right]$ is symmetric and doubly stochastic, ensuring that its eigenvalues are real and sorted in non-increasing order: $1=\lambda_1(Q) \geq \lambda_2(Q) \geq \cdots \geq \lambda_n(Q) \geq -1$. Additionally, we assume that $ \beta = \max \left\{\left|\lambda_2(Q)\right|,\left|\lambda_n(Q)\right|\right\} < 1 $, where $\beta$ represents the second largest magnitude among the eigenvalues of $Q$. 

\end{assumption}

\begin{remark}
    \cite{chen2021accelerating} The quantity $\beta \in(0,1)$ represents the connectivity of the topology. A smaller value of $\beta$ indicates a better-connected network, while a larger value of $\beta$ suggests a more poorly connected topology.
\end{remark}

Under the stated assumptions, we show that the total deviation of clients from mean is vanishing for both $A$ and $B$. Formally, we have the following lemma.
\begin{lemma}\label{lemma_first}
Under Assumptions \ref{ass:smoothness}, \ref{assm: bounded gradient}, \ref{assm: bounded matrices}, \ref{assm: mixing matrix}, and with $\eta=\frac{1}{KT^{\frac{1}{2}}}$ in \eqref{eq:updata}, the following holds  
\begin{gather}
\frac{1}{n}\sum\limits_{i=1}^{n}\mathbb{E}\left\|A_i^{(t)}-\bar{A}^{(t)}\right\|^2_F \le \tfrac{M_A}{T^{\frac{1}{2}}} + a_0(t) ,\quad
\frac{1}{n}\sum\limits_{i=1}^{n}\mathbb{E} \left \|B_i^{(t)}-\bar{B}^{(t)}\right \|^2_F \leq \tfrac{M_B}{T^{\frac{1}{2}}},
\end{gather}
where $M_A=\frac{2(1+\beta^2)\beta^2 G^2 C_B^2}{(1 - \beta^2)^2}$, $M_B=\frac{2(1+\beta^2)\beta^2 G^2 C_A^2}{(1 - \beta^2)^2}$, $a_0(t) = \frac{1}{n}\rho^{t} \mathbb{E}\|[A_i^{(0)}] \|^2 $, and $\rho=\frac{1+\beta^2}{2}$.
\end{lemma}

Next, we provide the main result of the paper.
\begin{theorem}\label{theorem}
Let Assumptions \ref{ass:smoothness}, \ref{assm: bounded gradient}, \ref{assm: bounded matrices}, and \ref{assm: mixing matrix} hold. Suppose in \eqref{eq:updata} we set $\eta=\frac{1}{KT^{\frac{1}{2}}}$. Then, after $T$ communication rounds, the following holds  
\begin{align}
    \frac{1}{T}\sum_{t=0}^{T-1} &\left(\mathbb{E}\left\|\nabla_A f(\bar{B}^{(t)}\bar{A}^{(t)})\right\|^2_F+\mathbb{E}\left\|\nabla_B f(\bar{B}^{(t)}\bar{A}^{(t)})\right\|^2_F\right) \le \mathcal{O}\left(\tfrac{\Delta_f+\Tilde{M}_A+\Tilde{M}_B}{T^{\frac{1}{2}}}+\tfrac{M+\hat{M}_A+\hat{M}_B}{T}\right.\nonumber\\
    &\hspace{5cm}+\left.\tfrac{(J_A+J_B)\sum_{i=1}^n\mathbb{E} \|A_i^{(0)}\|^2_F}{T(1-\beta^2)}+\tfrac{L(C_A^4+C_B^4)G^2}{T^{\frac{1}{2}}}\right),
\end{align}
where $ \Delta_f = \mathbb{E}f(\bar{B}^{(0)}\bar{A}^{(0)})-\mathbb{E}f(\bar{B}^{(T)}\bar{A}^{(T)}) $, $M = 2C_A^4 C_B^2 L^2 G^2 + 2C_A^2 G^4 $, and $\Tilde{M}_A$, $\Tilde{M}_B$, $\hat{M}_A$, $\hat{M}_B$, $J_A$, $J_B$ are variables that depend on $C_A$, $C_B$, $G$, $L$, $\beta$ and are defined in Lemma \ref{lemma:bounds}.
\end{theorem}
\textit{Proof.} The full proof is provided in Appendix \ref{convergence}, and a summary of the main ideas is presented below.

\begin{remark}
    Focusing on the dependency of the convergence rate on the number of communication rounds $T$, the above theorem demonstrates that we can achieve a convergence rate of $\mathcal{O}(\frac{1}{T^{1/2}})$ toward the stationary solution.

\end{remark}

\subsection{Proof Steps}\label{proof_steps}
We begin with two key lemmas for our proof. Lemma \ref{lemma_smooth} analyzes the smoothness of local objective functions with respect to $A$ and $B$ when the other variable is held fixed. Lemma  \ref{lemma_bounds} uses Lemma \ref{lemma_first} to further show that the deviation between the sum of local gradients and the global gradient is vanishing.
\begin{lemma}\label{lemma_smooth}
    Under Assumption \ref{ass:smoothness}, each local objective function $f_i$ is $LC_B^2$-smooth with respect to $A$ when $B$ is fixed, and $LC_A^2$-smooth with respect to $B$ when $A$ is fixed.
\end{lemma}

\begin{lemma}\label{lemma_bounds}
Under Assumptions \ref{ass:smoothness}, \ref{assm: bounded gradient}, \ref{assm: bounded matrices}, \ref{assm: mixing matrix}, we have the following bounds
\begin{align}
    \mathbb{E}\left\|g_
    A^{(t)+k}-\nabla_Af(\bar{B}^{(t)}\bar{A}^{(t)})\right\|^2_F \le K^2\eta^2\Tilde{M}_A+K^2\eta^2\hat{M}_A+J_A a_0(t)\nonumber,\\
    \mathbb{E}\left\|g_
    B^{(t)+k}-\nabla_Bf(\bar{B}^{(t)}\bar{A}^{(t)})\right\|^2_F \le K^2\eta^2\Tilde{M}_B+K^2\eta^2\hat{M}_B+J_B a_0(t),
\end{align}
where $g_A^{(t)+k}=\frac{1}{n} \sum_{i=1}^n \nabla_A f_i(W_i^{(t)+k})$ and $g_B^{(t)+k}=\frac{1}{n} \sum_{i=1}^n \nabla_B f_i(W_i^{(t)+k})$.
\end{lemma}
Before presenting the proof, we briefly outline the technique used. We begin by applying Lemma~\ref{lemma_smooth} to invoke the smoothness property with respect to \( A \) and \( B \) separately at appropriate points. When summing the resulting inequalities~\eqref{first_phrase} and~\eqref{second_phrase}, the additional cross terms, highlighted in blue, cancel out, allowing us to recover the desired objective. Note that the steps leading to inequalities marked as \( (a) \) are detailed in Appendix~\ref{convergence}.

\textbf{Proof of Theorem \ref{theorem}.} 
Using smoothness for $A$ from Lemma \ref{lemma_smooth}, and noticing that 
\begin{align}
    \bar{A}^{(t)} = \frac{1}{n}\sum_{i,j=1}^n q_{ij} \left(A_j^{(t-1)}- \eta\sum_{k=0}^{K-1}\nabla_A F_i(W_i^{(t-1)+k};\xi_i^{(t-1)+k})\right) = \bar{A}^{(t-1)} - \eta\sum_{k=0}^{K-1} \Tilde{g}_A^{(t-1)+k},
\end{align}
we can write
\begin{align}
    \mathbb{E}f(\bar{B}^{(t)}\bar{A}^{(t)})&\le \textcolor{blue}{\mathbb{E}f(\bar{B}^{(t)}\bar{A}^{(t-1)})} +\mathbb{E}\left\langle \nabla_A f(\bar{B}^{(t)}\bar{A}^{(t-1)}), \bar{A}^{(t)}-\bar{A}^{(t-1)}\right\rangle +\tfrac{LC_B^2}{2}\mathbb{E}\left\|\bar{A}^{(t)}-\bar{A}^{(t-1)}\right\|^2_F\nonumber\\
    &\hspace{-1.5cm}\stackrel{(a)}{\le} \mathbb{E}f(\bar{B}^{(t)}\bar{A}^{(t-1)}) +2K^3\eta^3M -\tfrac{K\eta }{2}\mathbb{E}\left\|\nabla_A f(\bar{B}^{(t-1)}\bar{A}^{(t-1)})\right\|^2_F+\tfrac{LC_B^4K^2G^2\eta^2}{2} \nonumber\\&
    \quad  + \tfrac{5\eta}{4}\sum_{k=0}^{K-1}\mathbb{E}\left\|g_A^{(t-1)+k}-\nabla_A f(\bar{B}^{(t-1)}\bar{A}^{(t-1)})\right\|^2_F\label{first_phrase}.
\end{align}
Using smoothness for $B$ from Lemma \ref{lemma_smooth}, we can write
\begin{align}
    \textcolor{blue}{\mathbb{E}}&\textcolor{blue}{f(\bar{B}^{(t)}\bar{A}^{(t-1)})}\le \mathbb{E}f(\bar{B}^{(t-1)}\bar{A}^{(t-1)})+\mathbb{E}\left\langle \nabla_B f(\bar{B}^{(t-1)}\bar{A}^{(t-1)}), \bar{B}^{(t)}-\bar{B}^{(t-1)}\right\rangle\nonumber\\
    &+ \tfrac{LC_A^2}{2}\mathbb{E}\left\|\bar{B}^{(t)}-\bar{B}^{(t-1)}\right\|^2_F
    \stackrel{(a)}{\le}\mathbb{E}f(\bar{B}^{(t-1)}\bar{A}^{(t-1)}) - \tfrac{K\eta}{2}\mathbb{E}\left\|\nabla_B f(\bar{B}^{(t-1)}\bar{A}^{(t-1)})\right\|^2_F \nonumber\\
    &\hspace{3cm}+ \tfrac{\eta}{2}\sum_{k=0}^{K-1}\mathbb{E} \left\| g_B^{(t-1)+k} -\nabla_B f(\bar{B}^{(t-1)}\bar{A}^{(t-1)})\right\|^2_F+ \tfrac{LC_A^4 K^2G^2\eta^2}{2}.\label{second_phrase}
\end{align}
Summing up \eqref{first_phrase} and \eqref{second_phrase}, using Lemma \ref{lemma_bounds}, and rearranging we obtain
\begin{align}
    \mathbb{E}&\left\|\nabla_A f(\bar{B}^{(t-1)}\bar{A}^{(t-1)})\right\|^2_F+\mathbb{E}\left\|\nabla_B f(\bar{B}^{(t-1)}\bar{A}^{(t-1)})\right\|^2_F
    \le \tfrac{2\left(\mathbb{E}f(\bar{B}^{(t-1)}\bar{A}^{(t-1)})-\mathbb{E}f(\bar{B}^{(t)}\bar{A}^{(t)})\right)}{K\eta}\nonumber\\
    &\hspace{0.8cm}+ \left(\tfrac{5\Tilde{M}_A+2\Tilde{M}_B}{2}\right)K^2\eta^2+\left(\tfrac{5\hat{M}_A+2\hat{M}_B}{2}+4M\right)K^2\eta^2+\left(\tfrac{5 J_A+2J_B}{2}\right) a_0(t-1)\nonumber\\
    &\hspace{1.1cm}+\left(C_A^4+C_B^4\right)LG^2K\eta.\label{eq::10}
\end{align}
By repeatedly applying \eqref{eq::10} for different values of $t$ and summing the results, we obtain
\begin{align}
     \frac{1}{T}\sum_{t=0}^{T-1}&\left(\mathbb{E}\left\|\nabla_A f(\bar{B}^{(t)}\bar{A}^{(t)})\right\|^2_F+\mathbb{E}\left\|\nabla_B f(\bar{B}^{(t)}\bar{A}^{(t)})\right\|^2_F\right)
    \le \tfrac{2\left(\mathbb{E}f(\bar{B}^{(0)}\bar{A}^{(0)})-\mathbb{E}f(\bar{B}^{(T)}\bar{A}^{(T)})\right)}{TK\eta}\nonumber\\
    &\quad+ \left(\tfrac{5\Tilde{M}_A+2\Tilde{M}_B}{2}\right)K^2\eta^2+\left(\tfrac{5\hat{M}_A+2\hat{M}_B}{2}+4M\right)K^2\eta^2+\left(\tfrac{5 J_A+2J_B}{2}\right) \tfrac{\sum_{t=0}^{T-1}a_0(t)}{T}\nonumber\\
    &\quad\quad+\left(C_A^4+C_B^4\right)LG^2K\eta.
\end{align}
Using that $ \sum_t a_0(t) \le \frac{\sum_{i=1}^n\mathbb{E} \|A_i^{(0)}\|^2_F}{n(1-\rho)} $ and setting $\eta = \tfrac{1}{KT^{\frac{1}{2}}}$, completes the proof.\hfill \ensuremath{\Box}

\section{Empirical Results}


\subsection{Experiments Setup}\label{exp:setup}
We conduct extensive experiments to evaluate the performance of the proposed algorithm across two language models. For the BERT-family models, we utilize RoBERTa-base~\cite{liu2019roberta}, while for large-scale models, we employ LLaMA-2-7B~\cite{touvron2023llama}. To evaluate \texttt{Dec-LoRA}, we consider two topologies: a Ring topology, where each client connects to two neighbors, and an Erd\H{o}s-R\'enyi (ER) topology. The mixing matrix for the ER topology is defined as $Q = I - \frac{2}{3 \lambda_{\max}(L)} L$, where $L$ is the Laplacian matrix of an ER graph with edge probability $p_c$. A larger $p_c$ results in a more connected graph. We have provided details regarding these topologies in Appendix \ref{network_topology}. We perform the experiments on NVIDIA A6000 and V100 GPUs.

\textbf{Comparative methods.}\label{others}
We compare our proposed \texttt{Dec-LoRA} method with three widely used PEFT approaches in a decentralized setting: Adapter~\cite{houlsby2019parameter} (\texttt{Dec-Adapter}), BitFit~\cite{zaken2021bitfit} (\texttt{Dec-BitFit}), and IA3~\cite{liu2022few} (\texttt{Dec-IA3}). These methods are implemented using the Hugging Face PEFT library~\cite{peft}. Additionally, we maintain the default hyperparameter settings for the baseline methods to ensure consistency and generalizability across all tasks.

\subsection{Performance on the BERT Family}\label{bert}
We conduct experiments using MRPC, SST-2, QNLI, QQP, and MNLI datasets from the Generalized Language Understanding Evaluation (GLUE) benchmark~\cite{wang2018glue}, utilizing the full training dataset for each task and reporting the best validation accuracy. Validation accuracies are calculated based on the averaged models of the clients at the end of each communication round. Additional details regarding the experiments conducted on the BERT family can be found in the Appendix \ref{app.A}.

\subsubsection{Comparative Analysis of Decentralized PEFT methods}\label{comparitive1}
In this section, we compare the convergence speeds and accuracies of \texttt{Dec-LoRA} with three other methods discussed earlier. Table \ref{tab:com} presents the results after $20$ iterations for experiments with $K=1$, and $10$ iterations for experiments with $K=5$. We set the rank to $16$ for \texttt{Dec-LoRA} and the bottleneck size to $64$ for \texttt{Dec-Adapter}. The experiments are conducted using ring and ER topologies with $10$ clients. As shown in the table, \texttt{Dec-LoRA} achieves the highest average accuracy among these methods, while maintaining a relatively low number of trainable parameters. Additionally, the convergence speed for the Ring topology with $K=1$ is depicted in Fig. \ref{fig:method2}. As illustrated, \texttt{Dec-LoRA} demonstrates faster convergence compared to the other methods across various tasks.
\begin{table*}[t]
\caption{A comparative analysis of various decentralized PEFT methods using the RoBERTa-Base model. Highest accuracy is highlighted in \textbf{bold}, and the second highest is \underline{underlined}.}
\label{tab:com}
\begin{center}
\resizebox{0.75\linewidth}{!}{
\begin{tabular}{clcccccccccc}
\toprule 
\multirow{2.5}{*}{} &
 \multirow{2.5}{*}{Method} &
 \multirow{2.5}{*}{\# Param.} &
 \multicolumn{2}{c}{QNLI} & \multicolumn{2}{c}{SST2} & \multicolumn{2}{c}{MNLI} &
 \multicolumn{2}{c}{QQP}&
 \multirow{2.5}{*}{Avg.}\\
 \cmidrule(lr){4-5} \cmidrule(lr){6-7} \cmidrule(lr){8-9} \cmidrule(lr){10-11} 
   & & &  Ring & ER & Ring & ER & Ring & ER & Ring & ER \\
\midrule
\multirow{4}{*}{\rotatebox{90}{$K=1$}}  &
\textbf{\texttt{Dec-LoRA}} & \underline{$0.60M$}  
&$\bm{90.99}$ & $\bm{90.81}$ & $\bm{93.81}$ & \underline{$93.92$} & \underline{$85.37$}& $\bm{85.74}$ & $\bm{88.19}$ & $\bm{88.01}$ & $\bm{89.61}$ \\
&\texttt{Dec-Adapter} & $2.95M$
&  \underline{$90.52$}& \underline{$90.54$} & \underline{$93.58$}& $\bm{94.38}$ &  $\bm{85.45}$& \underline{$84.92$} & \underline{$87.92$}& \underline{$87.81$} & \underline{$89.39$} \\
&\texttt{Dec-BitFit} & $\bm{0.10M}$
&$86.47$& $85.81$ & $92.43$& $92.78$ & $83.76$& $84.24$ & $82.39$ & $83.46$ & $86.42$\\
&\texttt{Dec-IA3} & $0.65M$
& $89.36$ & $89.05$ & $92.66$& $92.55$ & $83.61$ & $83.61$ & $85.86$& $85.53$ & $87.78$ \\
\midrule
\multirow{4}{*}{\rotatebox{90}{$K=5$}} &
\textbf{\texttt{Dec-LoRA}} & \underline{$0.60M$}  
& $\bm{91.23}$ & $\bm{91.63}$ & $\bm{94.61}$& $\underline{94.27}$ & $\bm{85.94}$ & $\bm{85.60}$ & $\bm{85.10}$& $\underline{86.76}$ & $\bm{89.39}$ \\
&\texttt{Dec-Adapter} & $2.95M$
& $\underline{90.72}$ & $\underline{90.08}$ & $\underline{93.69}$& $\bm{94.38}$ & $82.28$ & $83.65$ & $81.01$& $85.01$ & $87.60$ \\
&\texttt{Dec-BitFit} & $\bm{0.10M}$
& $88.28$ & $89.42$ & $93.35$& $93.12$ & $82.28$ & $82.50$ & $80.59$& $85.35$ & $86.86$\\
&\texttt{Dec-IA3} & $0.65M$ 
& $90.12$ & $89.97$ & $93.00$& $93.23$ & $\underline{84.89}$ & $\underline{84.50}$ & $\underline{84.02}$& $\bm{87.04}$ & $\underline{88.35}$ \\
\bottomrule
\end{tabular}
}
\end{center}
\end{table*}

    

\begin{figure}[t]
    \centering 
    \scriptsize
    \begin{subfigure}[b]{0.32\textwidth} 
        \includegraphics[width=\textwidth]{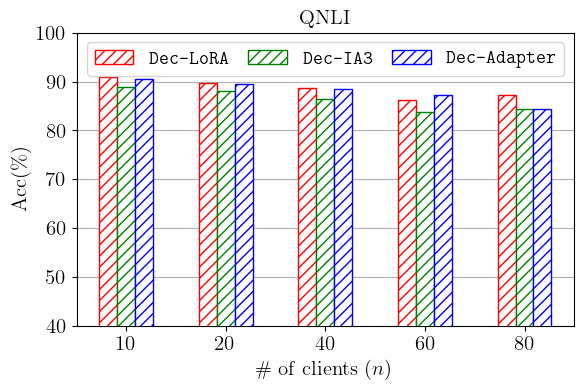}
        \caption{}
    \end{subfigure}
    \hspace{0.1cm}
    \begin{subfigure}[b]{0.32\textwidth} 
        \includegraphics[width=\textwidth]{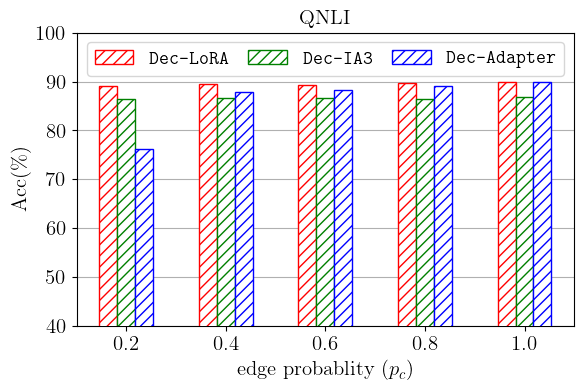}
        \caption{}
    \end{subfigure}
    \hspace{0.1cm}
    \begin{subfigure}[b]{0.32\textwidth} 
        \includegraphics[width=\textwidth]{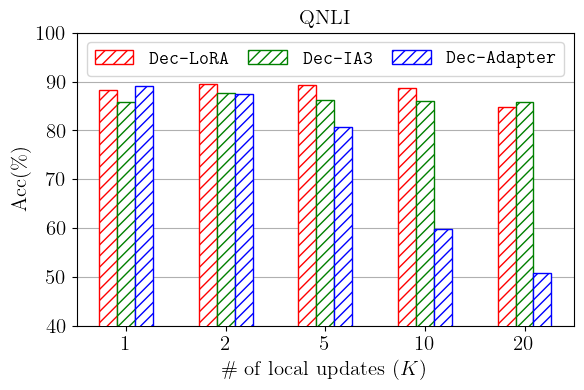}
        \caption{}
    \end{subfigure}


    \begin{subfigure}[b]{0.32\textwidth} 
        \includegraphics[width=\textwidth]{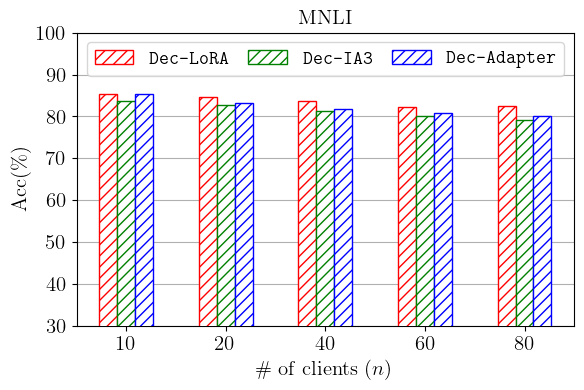}
        \caption{}
    \end{subfigure}
    \hspace{0.1cm}
    \begin{subfigure}[b]{0.32\textwidth} 
        \includegraphics[width=\textwidth]{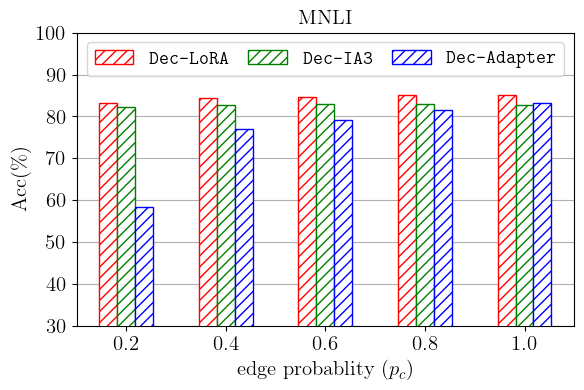}
        \caption{}
    \end{subfigure}
    \hspace{0.1cm}
    \begin{subfigure}[b]{0.32\textwidth} 
        \includegraphics[width=\textwidth]{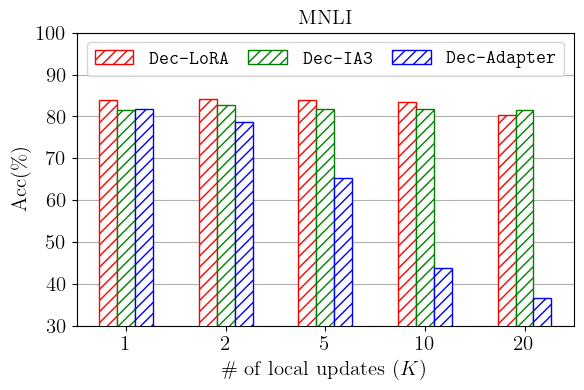}
        \caption{}
    \end{subfigure}
    
    \caption{(a) and (d): Effect of the number of clients on accuracy for the Ring topology with $K=1$.  
(b) and (e): Effect of edge probability in the Erd\H{o}s-R\'enyi topology on accuracy for $K=5$.  
(c) and (f): Effect of the number of local updates ($K$) on accuracy for the Ring topology.
}
    \label{fig2}
\end{figure}
\subsubsection{Impact of Number of Clients, Edge Probabilities, and Number of Local Updates}\label{comparitive2}

To illustrate the impact of various parameters during the fine-tuning process, we present results for three methods on the QNLI and MNLI datasets in Fig.~\ref{fig2}. As shown, \texttt{Dec-LoRA} outperforms the baselines across most settings. The detailed results are as follows. \textbf{(1) Fig.~\ref{fig2} (a) and (d):} These plots show the effect of the number of clients on accuracy for the Ring topology with \(T=20\) and \(K=1\). As expected, the accuracy generally decreases as the number of clients increases across different tasks. \textbf{(2) Fig.~\ref{fig2} (b) and (e):} These plots highlight the influence of edge probability in the ER topology on accuracy, with parameters set to \(N=30\), \(T=5\), and \(K=5\). As demonstrated, a more connected network, characterized by a higher edge probability (\(p_c\)), leads to improved accuracy. \textbf{(3) Fig.~\ref{fig2} (c) and (f):} These figures show the effect of the number of local updates on accuracy for the Ring topology with \(N=30\). In these cases, \(K \times T = 20\) for all experiments. While an increase in the number of local updates enhances communication efficiency, it results in lower accuracy when the total gradient computation remains constant. We validate these empirical results with our theoretical findings in Appendix \ref{app:impact}.

\subsubsection{Comparison of \texttt{Dec-LoRA} with centralized LoRA}\label{comparison_center}
\begin{figure}[t]
    \centering 
    \scriptsize
    \begin{subfigure}[b]{0.23\textwidth} 
        \includegraphics[width=\textwidth]{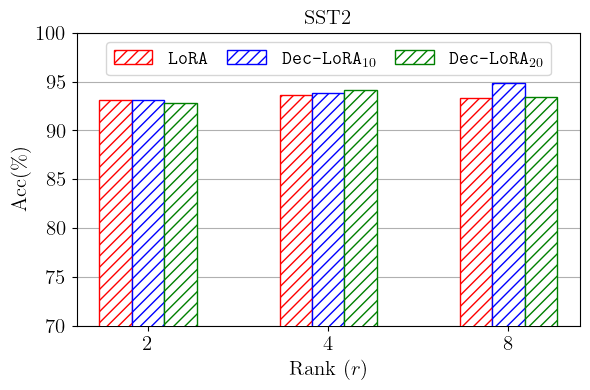}
    \end{subfigure}
    \hspace{0.1cm}
    \begin{subfigure}[b]{0.23\textwidth} 
        \includegraphics[width=\textwidth]{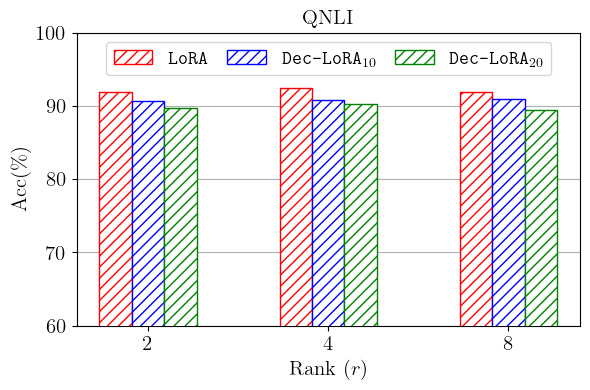}
    \end{subfigure}
    \hspace{0.1cm}
    \begin{subfigure}[b]{0.23\textwidth} 
        \includegraphics[width=\textwidth]{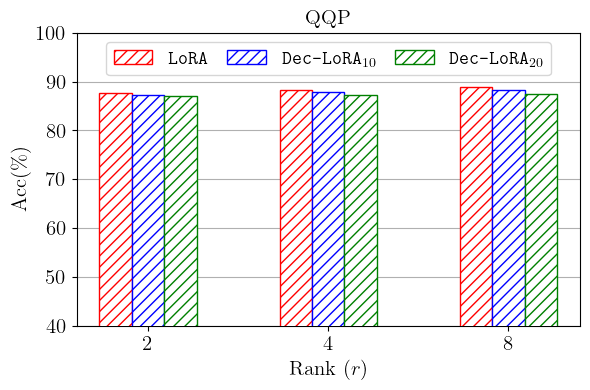}
    \end{subfigure}
\hspace{0.1cm}
    \begin{subfigure}[b]{0.23\textwidth} 
        \includegraphics[width=\textwidth]{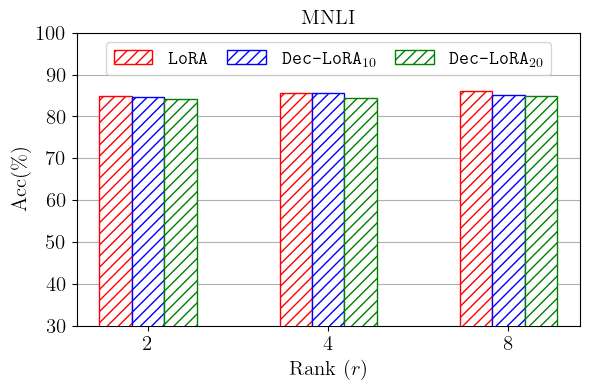}
    \end{subfigure}

    \caption{A comparative analysis of centralized LoRA and \texttt{Dec-LoRA} with $10$ and $20$ clients under different ranks, using the RoBERTa-base model.
}
    \label{fig_bar}
\end{figure}

We provide a comparative analysis of centralized and decentralized LoRA with $10$ and $20$ clients across various ranks for the Ring topology, evaluated on four datasets, as presented in Fig.~\ref{fig_bar}. The results, obtained after $100$ communication rounds, indicate that \texttt{Dec-LoRA} achieves accuracy levels comparable to centralized LoRA fine-tuning, highlighting its viability as an effective solution for decentralized settings.

\begin{table*}[t]
\caption{Left half: Performance analysis of \texttt{Dec-LoRA} with 4-bit quantization for $10$ clients across different ranks. Right half: Performance analysis of \texttt{Dec-LoRA} under data heterogeneity with $3$ clients across different ranks. }
\label{tab:quantization}
\begin{center}
\resizebox{\linewidth}{!}{
\begin{tabular}{lcccccccc|cccccccc}
\toprule 
 \multirow{2}{*}{Method (Rank)} & \multicolumn{2}{c}{QNLI} & \multicolumn{2}{c}{SST2} & \multicolumn{2}{c}{MRPC} &
 \multicolumn{2}{c|}{QQP} &
 \multicolumn{2}{c}{QNLI} & \multicolumn{2}{c}{SST2} & \multicolumn{2}{c}{MNLI} &
 \multicolumn{2}{c}{QQP} \\
 \cmidrule(lr){2-3} \cmidrule(lr){4-5} \cmidrule(lr){6-7} \cmidrule(lr){8-9} 
 \cmidrule(lr){10-11} \cmidrule(lr){12-13} \cmidrule(lr){14-15} \cmidrule(lr){16-17}
 & Full & 4-bit & Full & 4-bit & Full & 4-bit & Full & 4-bit & i.i.d. & non-i.i.d. & i.i.d. & non-i.i.d. & i.i.d. & non-i.i.d. & i.i.d. & non-i.i.d.\\
\midrule
\texttt{Dec-LoRA} ($2$) & $90.65$ & $90.35$ & $93.12$ & $94.38$ & $89.31$ & $89.53$ & $87.25$ & $87.32$ & $90.44$ & $89.99$ & $94.84$ & $94.27$ & $85.63$ & $85.39 $& $88.25$ & $86.99$ \\
\texttt{Dec-LoRA} ($4$) & $90.85$ & $91.01$ & $93.81$ & $93.69$ & $89.20$ & $88.97$ &  $87.89$ & $87.65$ & $91.18$ & $90.66$ & $95.18$ &$ 94.04$ & $85.71$ & $85.58$ & $88.70$ & $88.01$\\
\texttt{Dec-LoRA} ($8$) & $90.88$ & $91.16$ & $94.84$ & $93.69$ & $89.16$ & $91.45$ & $88.23$ & $88.42$ & $91.31$ & $89.90$ & $94.61$ & $94.72$ & $86.23$ & $84.15$ & $88.89$ & $88.24$ \\
\midrule\rowcolor{blue!10} 
Avg. & $90.79$ & $90.84$ & $93.92$ & $93.92$ & $89.22$ & $89.98$ & $87.79$ & $87.80$ & $90.98$ & $90.18$ & $94.88$ & $94.34$ & $85.86$ & $85.04$ & $88.61$ & $87.75$\\
\bottomrule
\end{tabular}
}
\end{center}
\end{table*}

\subsubsection{\texttt{Dec-LoRA} with Quantization}\label{comparison_quan}
In this section, we evaluate the use of LoRA with 4-bit quantization for the pretrained model (QLoRA) \cite{dettmers2024qlora} in a decentralized setting. Specifically, QLoRA leverages $4$-bit quantization to compress the base model, making it much more memory efficient, while still allowing for fine-tuning using trainable LoRA adapters. This technique is particularly suited for decentralized environments where computing resources are often limited, as it enables efficient training of large models on standard GPUs.

For our experiments, we consider a decentralized setup with $10$ clients arranged in a Ring topology. The results, presented on the left side of Table~\ref{tab:quantization}, show that \texttt{Dec-LoRA} with $4$-bit quantization of the pretrained model performs nearly identically to the regular \texttt{Dec-LoRA}. This demonstrates its potential to significantly reduce memory usage in decentralized settings.

\subsubsection{\texttt{Dec-LoRA} under Data Heterogeneity}\label{hetero}
Data heterogeneity occurs when the training data is not identically and independently distributed across clients (non-i.i.d.), causing local models on individual clients to deviate from the global model's optimal state, which can result in slower convergence~\cite{hsieh2020non,li2020federated}. 

In this section, we assess the performance of \texttt{Dec-LoRA} under the condition of data heterogeneity using three clients, following a setup similar to that in~\cite{sun2024improving}. For the heterogeneous setting, we partition the data based on class labels. For binary classification tasks, the data is split as $[0.15, 0.85]$, $[0.85, 0.15]$, and $[0.5, 0.5]$, while for three-class classification tasks, the splits are $[0.6, 0.2, 0.2]$, $[0.2, 0.6, 0.2]$, and $[0.2, 0.2, 0.6]$. The results are presented on the right side of Table~\ref{tab:quantization}. As observed, there is a slight drop in performance under the non-i.i.d. setting.
A more detailed discussion of this phenomenon can be found in Section~\ref{limitation}.

\subsection{Performance on the
Large-Scale Language Models}\label{large}

We evaluate performance using SuperGLUE tasks~\cite{wang2019superglue} and question-answering generation tasks, including SQuAD~\cite{rajpurkar2016squad} and DROP~\cite{dua2019drop}. For each task, we randomly select $1000$ samples for training and $1000$ samples for validation, reporting the best validation accuracy. Additional details regarding the experiments are provided in Appendix \ref{app.B}. Table~\ref{tab:generation} presents the results for LoRA and \texttt{Dec-LoRA} implemented under a Ring topology with $10$ clients and $3$ local updates, utilizing the LLaMA-2-7B model. As shown, for larger models, \texttt{Dec-LoRA} performs comparably to centralized LoRA on most tasks, indicating its effectiveness in decentralized environments. Additional experiments conducted on LLaMA-2-13B and OPT-2.7B~\cite{zhang2022opt} are presented in Table \ref{tab:generation2} in the Appendix.

\begin{table*}[t]
\caption{A comparative analysis of centralized LoRA and \texttt{Dec-LoRA} with $10$ clients under different ranks, using the LLaMA-2-7B model.}
\label{tab:generation}
\begin{center}
\resizebox{\linewidth}{!}{
\begin{tabular}{cccccccccccccc}
\toprule 
\multirow{4}{*}{Rank} & 
\multirow{4}{*}{\# Param.} &
\multicolumn{4}{c}{Classfication} & \multicolumn{4}{c}{Multiple Choice} & \multicolumn{4}{c}{Generation}\\
\cmidrule(lr){3-6} \cmidrule(lr){7-10} \cmidrule(lr){11-14}
  & &  \multicolumn{2}{c}{WIC} & \multicolumn{2}{c}{BoolQ} &
 \multicolumn{2}{c}{COPA} &
 \multicolumn{2}{c}{ReCoRD} &
 \multicolumn{2}{c}{SQuAD} &
 \multicolumn{2}{c}{DROP}\\
 \cmidrule(lr){3-4} \cmidrule(lr){5-6} \cmidrule(lr){7-8} \cmidrule(lr){9-10} \cmidrule(lr){11-12} \cmidrule(lr){13-14} 
 & & \texttt{LoRA} & \texttt{Dec-LoRA}$_{10}$  & \texttt{LoRA} & \texttt{Dec-LoRA}$_{10}$  & \texttt{LoRA} & \texttt{Dec-LoRA}$_{10}$  & \texttt{LoRA} & \texttt{Dec-LoRA}$_{10}$  & \texttt{LoRA} & \texttt{Dec-LoRA}$_{10}$  & \texttt{LoRA} & \texttt{Dec-LoRA}$_{10}$ \\
\midrule
$2$ & $1.05M$ & $73.20$ & $72.57$ & $85.9$ & $84.0$ & $87$ & $87$ & $82.4$ & $81.1$ & $89.76$ & $89.39$ & $48.32$ & $44.35$  \\
$4$ & $2.10M$ & $74.61$ & $72.26$ & $85.2$ & $83.5$ & $85$ & $87$ & $81.1$ & $81.2$ & $89.79$ & $89.79$ & $46.56$ & $44.97$  \\
$8$ & $4.19M$ & $73.04$ & $69.44$ & $85.4$ & $83.7$ & $85$ & $89$ & $81.0$ & $81.3$ & $90.11$ & $89.93$ & $47.59$ & $44.99$  \\
\midrule\rowcolor{blue!10} 
Avg. & $2.44M$ & $73.62$ & $71.42$ & $85.5$ & $83.7$  & $86$ & $88$ & $81.5$ & $81.2$ & $89.89$ & $89.70$ & $47.49$ & $44.77$ \\
\bottomrule
\end{tabular}
}
\end{center}
\end{table*}

\section{Conclusion}

In this work, we introduced \texttt{Dec-LoRA}, a method for decentralized federated fine-tuning of LLMs using LoRA. By removing the need for a central server, Dec-LoRA allows efficient and federated model adaptation in distributed settings while preserving data privacy. Extensive experiments on BERT and LLaMA-2 family models demonstrate that \texttt{Dec-LoRA} outperforms several popular PEFT approaches and achieves performance comparable to centralized LoRA, even under challenging conditions such as data heterogeneity and quantization constraints. We further provide a theoretical analysis demonstrating that, under standard assumptions such as the smoothness of the loss function and bounded gradients, our proposed \texttt{Dec-LoRA} method can converge to a stationary solution.
 These findings highlight the potential of decentralized federated fine-tuning as a viable alternative to traditional federated approaches, opening new opportunities for future research in collaborative, serverless adaptation of LLMs.

\bibliographystyle{ieeetr} 
\bibliography{main.bib}

\pagebreak
\appendix


\section{Pipelines of the Proposed Algorithm}\label{pipeline}
For a better understanding of the steps of the designed algorithm, we present \texttt{Dec-LoRA}  in Algorithm \ref{al.1}.
\begin{algorithm}[t]
	\caption{\texttt{Dec-LoRA}} \label{al.1}
 {\small
	\begin{algorithmic}[1]
            \State\textbf{Input:} Step size $\eta$, number of communication rounds $T$, number of local updates $K$, number of clients $n$, mixing matrix $Q$.
            \State \textbf{Initialize:} $\forall c_i \in \mathcal{C}$, $A_i^{(0)} = A_0 \sim \mathcal{N}(0, \sigma^2)$, $B_i^{(0)} = 0$.\Comment{Standard LoRA initialization \cite{hu2021lora}}
            \For {communication round $ t\leftarrow1 $ to $ T $}
		\For {clients $ c_i \in \mathcal{C} $ in parallel} 
                \For 
                {local update $ k\leftarrow1 $ to $ K $}
                \State $A_i^{(t)+k+1}= A_i^{(t)+k}-\eta \Tilde{\nabla}_A f_i(W_i^{(t)+k} )$ \Comment{Local training on $A$}
                \State $B_i^{(t)+k+1} = B_i^{(t)+k}-\eta \Tilde{\nabla}_B f_i(W_i^{(t)+k})$
                \Comment{Local training on $B$}
                \EndFor
                \State Client $c_i$ sends $ A_i^{(t)+K} $ and $B_i^{(t)+K}$ to their neighbors 
                \EndFor
                \State At Client $c_i$:  
                $ \text{  }A_i^{(t+1)}=\sum_j q_{i j} A_j^{(t)+K}$\\$
    \quad\quad\quad\quad\quad\quad\quad B_i^{(t+1)}=\sum_j q_{i j} B_j^{(t)+K}$ \Comment{Aggregation using mixing matrix $Q$}
       
				\EndFor
                \State Randomly draw $\hat{A}, \hat{B}$ from $\{\bar{A}^{(t)},\bar{B}^{(t)}\}_{t=1}^T$ at uniform.
                \State \textbf{Return: } $\hat{A}, \hat{B}$.
	\end{algorithmic} 
 }
\end{algorithm}

\section{Related Work}
\subsection{Parameter Efficient Fine-Tuning on LLMs}
LLMs such as GPT-4~\cite{achiam2023gpt}, LLaMA~\cite{touvron2023llama}, and BERT~\cite{devlin2018bert} have achieved remarkable performance across various tasks like translation and summarization~\cite{bommasani2021opportunities} due to architectures like Transformers~\cite{vaswani2017attention}. However, these models typically contain billions of trainable parameters, making full fine-tuning (FFT) computationally expensive and inefficient, particularly for task-specific adaptations. To address this, PEFT methods have been introduced, enabling adaptation with significantly fewer trainable parameters while maintaining performance close to FFT. PEFT methods can be generally divided into three categories~\cite{han2024parameter}. \textit{Additive}  introduces a small set of trainable parameters while keeping the original model frozen, as seen in Serial Adapter~\cite{houlsby2019parameter}, Parallel Adapter~\cite{he2021towards}, Prefix-Tuning~\cite{li2021prefix}, and Prompt-Tuning~\cite{lester2021power}. \textit{Selective} PEFT fine-tunes only a subset of existing model parameters, with techniques like BitFit~\cite{zaken2021bitfit} and PaFi~\cite{liao2023parameter}. \textit{Reparameterized} PEFT introduces a low-rank parameterization of pre-trained weights for training, with methods such as LoRA~\cite{hu2021lora} and DoRA~\cite{liu2024dora}. \textit{Among these, LoRA stands out for its efficiency, effectiveness, and adaptability, making it a compelling choice for fine-tuning LLMs. In this work, we specifically focus on the decentralization of LoRA.}

\subsection{PEFT in Federated Setting}
In their studies, \cite{zhang2023fedpetuning,fan2023fate} evaluate and compare various PEFT methods, including Adapters, LoRA, Prompt Tuning, and BitFit in FL. Several adaptations of LoRA have been introduced to enhance its efficiency in highly heterogeneous federated settings. For instance, SLoRA~\cite{babakniya2023slora,yan2024federa} modifies the initialization process to better handle data heterogeneity, while HetLoRA~\cite{cho2023heterogeneous} and FlexLoRA~\cite{bai2024federated} dynamically adjust LoRA ranks per client to account for system heterogeneity. More recently, FLoRA~\cite{wang2024flora} introduces slack matrices $A$ and $B$ for all clients and multiplies the resulting matrices to mitigate interference caused by the FedAvg algorithm.
To reduce communication overhead in federated LoRA, \cite{kuo2024federated} propose sparse fine-tuning techniques. Meanwhile, FFA-LoRA~\cite{sun2024improving} and RoLoRA~\cite{chenrobust} aim to enhance model accuracy in heterogeneous environments while minimizing the number of trainable parameters. Additionally, FedTT~\cite{ghiasvand2024communicationllm} integrates tensorized adapters for federated fine-tuning, significantly reducing trainable parameters and improving communication efficiency. \textit{Although extensive research has explored PEFT methods, particularly LoRA in centralized FL, no study has examined their performance in a fully decentralized setting without a central server, despite its relevance to many real-world applications.}

\subsection{Decentralized Optimization/Learning}
The exploration of decentralized optimization techniques dates back to at least the 1980s~\cite{tsitsiklis1984problems}. These algorithms, often called \textit{gossip algorithms}~\cite{kempe2003gossip,boyd2006randomized}, are characterized by the absence of a central authority for spreading information. Instead, information propagates through the network, similar to how gossip spreads along the edges defined by the communication graph. Among the most commonly used methods in decentralized optimization are those based on (sub)gradient descent~\cite{nedic2009distributed,johansson2010randomized}.
\\~\\
Decentralized optimization has recently facilitated the growth of decentralized learning, which has found applications in various domains, including autonomous vehicles~\cite{chellapandi2023federated}, healthcare systems~\cite{warnat2021swarm}, industrial IoT environments~\cite{qiu2022decentralized, hexmoor2024iot,ghajari2025network}, and social networks~\cite{he2022spreadgnn}.
In particular, decentralization has demonstrated exceptional effectiveness in LLM-based multi-agent systems, enabling scalable and robust collaboration among distributed agents~\cite{guo2024large,chen2024scalable}. \textit{Although PEFT methods, such as LoRA, can be beneficial for decentralized training of LLMs due to the large scale of these models, there is a lack of analysis on the use of such methods in decentralized scenarios. This paper aims to address this gap.}


\subsection{Theoretical LoRA}\label{theory_lora}
Strong theoretical guarantees for LoRA have yet to be discovered, but several important directions have emerged in the literature. \cite{jang2024lora} proves that LoRA with rank \( \mathcal{O}(\sqrt{N}) \), trained with gradient descent on \( N \) data points, can converge to the global minimizer under the neural tangent kernel (NTK) regime \cite{jacot2018neural}. \cite{zhuasymmetry} provides generalization bounds for LoRA with \( \sigma \)-sub-Gaussian loss functions. From a more optimization-oriented perspective, \cite{dayi2024gradient} explores the SGD dynamics of LoRA. However, their analysis is limited to 2-layer neural networks, and the fine-tuning target is rank-1 LoRA. Some works have borrowed ideas from matrix factorization~\cite{soltanolkotabi2023implicit} and adapted LoRA to provide theoretical guarantees on convergence to stationary solutions. For example, \cite{li2024crucial} introduces Nyström initialization for LoRA, while \cite{zhang2025one} introduces spectral initialization. However, the former assumes linear loss functions, and the latter considers either linear or 1-layer nonlinear loss functions with ReLU activation. \cite{malinovsky2024randomized} modifies LoRA by randomly generating and fixing \( A \) or \( B \) at the beginning of the training. Although they provide strong guarantees to stationary solutions for non-convex functions under reasonable assumptions, such as smoothness, their method does not yield competitive empirical results, especially when compared to the original LoRA.
\begin{table}[t]
	\centering
	\caption{Comparative summary of convergence rates from existing LoRA-related works in both centralized and federated settings.}
	\label{tab:rates}
	\resizebox{0.9\textwidth}{!}{%
		\begin{tabular}{@{}ccccccc@{}}
			\toprule
			Method & Federated & Rate & Assumption& Stochastic & FL Topology & Note                     \\ \midrule
                        \texttt{RAC-LoRA}~\cite{malinovsky2024randomized} & $\times$ & $ \mathcal{O}(1 / \sqrt{T})$ & non-convex & $\checkmark$ & N/A & - \\ 
                        \texttt{LoRA-One}~\cite{zhang2025one}& $\times$ & Linear Conv. & 1-layer non-linear loss & $\times$ & N/A & - \\
                        \midrule
                       \texttt{FLoRA}~\cite{wang2024flora}  &  $\checkmark$ & $\mathcal{O}(1 / T)$ & strongly-convex &  $\checkmark$ &Centralized & Provides no proof\\
                       \texttt{FFA-LoRA}~\cite{sun2024improving} & $\checkmark$ & N/A & N/A & N/A & Centralized & -\\
                       \texttt{FedSA-LoRA}~\cite{guo2024selective}  & $\checkmark$ &  $ \mathcal{O}(1 / \sqrt{T})$&non-convex &$\checkmark$ & Centralized & Week metric (described in \ref{theory_lora})\\
                       \texttt{Fed-RAC-LoRA}\cite{malinovsky2024randomized} & $\checkmark$ & $ \mathcal{O}(1 / \sqrt{T})$ & non-convex & $\checkmark$ & Centralized & No empirical results \\
                       \midrule
                       \texttt{Dec-LoRA} (Ours) & $\checkmark$ & $ \mathcal{O}(1 / \sqrt{T})$ & non-convex & $\checkmark$ &Decentralized & Strong metric (described in \ref{theory_lora})\\ 
			\bottomrule
		\end{tabular}%
	}
\end{table}

In federated learning, providing theoretical guarantees is even more challenging since \( A \)-updates are aggregated separately from \( B \)-updates in the server, whereas the "mathematically correct" update would be to average the products \( BA \) \cite{sun2024improving}. To tackle this issue, \cite{guo2024selective} proposes the modified and personalized federated learning version of LoRA, where the \( A \) matrices are trained globally among the clients, while the \( B \) matrices are trained locally. It further provides the following theoretical guarantee for non-convex and smooth functions:
\begin{align}
    \frac{1}{n T} \sum_{i=1}^n \sum_{t=0}^{T-1} \mathbb{E}\left\|\nabla_W f_i\left(B_i^{(t)}A_i^{(t)}\right)\right\|_F^2 \leq \mathcal{O}\left(\frac{1}{\sqrt{T}}\right).
\end{align}
Although \cite{guo2024selective} presents promising results, it is important to note that the guarantee provided is not very strong, as it provides no information about the quality of the solution for the global objective function \( f = \frac{1}{n}\sum_{i}^n f_i \) in \eqref{lora_equation}; instead, it identifies stationary points for the local objective functions.
What we expect, under the assumption of non-convexity of the objective functions in federated learning, is to find a stationary point of the global objective function, i.e., showing that the following metric converges to zero as $T \rightarrow \infty$ \cite{fallah2020personalized,wang2020tackling}:\cite{fallah2020personalized,wang2020tackling}:
\begin{align}
    \frac{1}{T}\sum_{t=0}^{T-1}\mathbb{E}\left\| \frac{1}{n}\sum_{i=1}^n \nabla_W f_i\left(\frac{1}{n }\sum_{i=1}^n B_i^{(t)}A_i^{(t)}\right)\right\|_F^2.\label{federated_bound:1}
\end{align}
where \( W = W_0 + BA \). As mentioned earlier, deriving upper bounds for~\eqref{federated_bound:1} is challenging due to the separate updating and aggregation of \( A \) and \( B \). To address this, we introduce a more tractable and practically meaningful metric for federated LoRA-based algorithms, motivated by the structure of typical LoRA implementations, defined as follows:
{\small
\begin{align}
\frac{1}{T}\sum_{t=0}^{T-1} &\mathbb{E}\left\| \frac{1}{n}\sum_{i=1}^n \nabla_{(A,B)} f_i\left(\frac{1}{n}\sum_{i=1}^n B_i^{(t)}\frac{1}{n}\sum_{i=1}^n A_i^{(t)}\right)\right\|_F^2 \nonumber\\
    =\frac{1}{T}\sum_{t=0}^{T-1} &\left(\mathbb{E}\left\| \frac{1}{n}\sum_{i=1}^n \nabla_A f_i\left(\frac{1}{n}\sum_{i=1}^n B_i^{(t)}\frac{1}{n}\sum_{i=1}^n A_i^{(t)}\right)\right\|^2+\mathbb{E}\left\| \frac{1}{n}\sum_{i=1}^n\nabla_Bf_i\left(\frac{1}{n}\sum_{i=1}^n B_i^{(t)}\frac{1}{n}\sum_{i=1}^nA_i^{(t)}\right)\right\|^2\right).
\end{align}}

This metric offers two key advantages: (1) it directly captures the stationarity condition with respect to the optimization variables \( (A, B) \), which represent the true degrees of freedom in low-rank adaptation; and (2) since the updates to \( A \) and \( B \) are aggregated separately, the search for a stationary solution is conducted over points of the form \( \bar{B}^{(t)}\bar{A}^{(t)} \), rather than the average product \( \frac{1}{n} \sum_{i=1}^n B_i^{(t)} A_i^{(t)} \).

Through a detailed convergence analysis, we prove that our decentralized LoRA algorithm converges to a stationary point with respect to this metric. 

We provide a comparative summary of convergence rates from existing LoRA-related works in both centralized and federated settings in Table \ref{tab:rates}.

\section{Bandwidth and Latency Advantage of \texttt{Dec-LoRA}}
In this section, we compare our proposed decentralized FL algorithm with the centralized FL counterpart in terms of bandwidth and latency.
\begin{itemize}
    \item \textbf{Bandwidth advantage:} In a LoRA-based approach, the data communicated per client per round consists of the LoRA matrices, with a size of $O((d_1+d_2)r)$. The key difference lies in how this data flows through the network:
    \begin{itemize}
        \item In \textit{Centralized FL}: All $n$ participating clients send their $O((d_1+d_2)r)$ updates to a single central server. This means the server must have an incoming bandwidth capable of handling $n * O((d_1+d_2)r)$ data per round. It then performs the aggregation and broadcasts the new model back out. The server's computational and communication load scales linearly with the number of clients $n$, making it a severe \textit{bottleneck} and limiting the scalability of the entire system.
        \item In \textit{Decentralized FL} (Ours): There is no central server. Each client sends its $O((d_1+d_2)r)$ update only to its immediate neighbors. For a client in a sparse network (like a Ring or a typical ER graph), the number of neighbors is a small constant, $d_n$, which is independent of the total number of clients $N$. Therefore, each node's communication load is only $d_n * O((d_1+d_2)r)$, which does not grow as the network size $n$ increases.
    \end{itemize}
    \item \textbf{Latency advantage}: Centralized FL requires a full round-trip communication (clients to server and server to clients), with latency dictated by the server's aggregation time. Decentralized communication happens in parallel between neighbors, leading to potentially much faster rounds.
\end{itemize}

\section{Additional Details for Empirical Results}\label{details_for_empirical}
\subsection{Ablation Study on Network Topology and Data Heterogeneity}
In this section, we conduct additional experiments that examine the effect of data heterogeneity under varying degrees of non-iidness, different network topologies, and two different numbers of clients ($30$ and $100$). These results are summarized in Table \ref{tab:topology}.

In all experiments, we fix the number of local updates to $2$. We simulate non-IID data distributions using a Dirichlet distribution over the class label proportions. Specifically, we consider three data partitions: Split-1 corresponds to an IID setting, Split-2 corresponds to a non-IID distribution with Dirichlet parameter $\alpha = 1$ (moderate non-iid), and Split-3 corresponds to a more heterogeneous distribution with Dirichlet parameter $\alpha = 0.5$ (severe non-iid).

\begin{table*}
\begin{center}
\caption{}
\label{tab:topology}
\resizebox{0.9\linewidth}{!}{
\begin{tabular}{clcccccccccccc}
\toprule
\multirow{2.5}{*}{} & \multirow{2.5}{*}{\textbf{Method (Topology)}} & \multicolumn{3}{c}{\textbf{QNLI}} & \multicolumn{3}{c}{\textbf{SST2}} & \multicolumn{3}{c}{\textbf{MNLI}} & \multicolumn{3}{c}{\textbf{Average}} \\
\cmidrule(lr){3-5} \cmidrule(lr){6-8} \cmidrule(lr){9-11} \cmidrule(lr){12-14}
& & Split-1 & Split-2 & Split-3 & Split-1 & Split-2 & Split-3 & Split-1 & Split-2 & Split-3 & Split-1 & Split-2 & Split-3 \\
\midrule
\multirow{10}{*}{\rotatebox{90}{$n=30$}} &
\texttt{Dec-LoRA} (complete) & 89.35 & 89.35 & 87.57 & 93.35 & 92.32 & 92.78 & 84.89 & 84.76 & 84.18 & 89.20 & 88.81 & 88.18 \\
&\texttt{Dec-LoRA} (ring) & 88.56 & 87.59 & 86.58 & 92.78 & 92.32 & 92.43 & 83.33 & 82.73 & 81.11 & 88.22 & 87.55 & 86.71 \\
&\texttt{Dec-LoRA} (ER-$p_c=0.6$) & 88.49 & 88.21 & 87.33 & 92.32 & 92.21 & 91.74 & 83.77 & 83.13 & 82.10 & 88.19 & 87.85 & 87.06 \\
&\texttt{Dec-LoRA} (ER-$p_c=0.2$) & 88.14 & 87.17 & 85.81 & 92.55 & 92.43 & 91.74 & 82.99 & 81.69 & 79.95 & 87.89 & 87.10 & 85.83 \\
&\texttt{Dec-LoRA} (exponential graph) & 89.35 & 89.05 & 87.39 & 93.00 & 92.66 & 93.12 & 84.67 & 84.54 & 84.23 & 89.01 & 88.75 & 88.25 \\
\cline{2-14}
&\texttt{Dec-Adapters} (complete) & 90.28 & 89.99 & 88.98 & 93.69 & 93.34 & 93.12 & 85.20 & 84.73 & 84.08 & 89.72 & 89.35 & 88.73 \\
&\texttt{Dec-Adapters} (ring) & 84.99 & 74.15 & 79.56 & 92.89 & 91.86 & 92.43 & 74.75 & 57.15 & 41.83 & 84.21 & 74.39 & 71.47 \\
&\texttt{Dec-Adapters} (ER-$p_c=0.6$) & 88.38 & 87.09 & 85.63 & 93.12 & 92.78 & 92.20 & 81.63 & 78.96 & 74.95 & 87.71 & 86.28 & 84.26 \\
&\texttt{Dec-Adapters} (ER-$p_c=0.2$) & 78.86 & 67.69 & 87.09 & 91.97 & 91.63 & 91.17 & 67.08 & 44.38 & 37.30 & 79.30 & 67.90 & 71.85 \\
&\texttt{Dec-Adapters} (exponential graph) & 90.13 & 90.04 & 89.55 & 93.46 & 93.12 & 93.81 & 84.90 & 84.32 & 83.44 & 89.50 & 89.16 & 88.94 \\
\midrule
\multirow{10}{*}{\rotatebox{90}{$n=100$}} &
\texttt{Dec-LoRA} (complete) & 85.47 & 85.03 & 84.61 & 91.51 & 91.29 & 91.63 & 82.53 & 82.02 & 81.12 & 86.50 & 86.11 & 85.79 \\
&\texttt{Dec-LoRA} (ring) & 85.03 & 84.59 & 81.38 & 91.51 & 91.28 & 90.94 & 80.41 & 79.28 & 77.60 & 85.65 & 85.05 & 83.31 \\
&\texttt{Dec-LoRA} (ER-$p_c=0.6$) & 84.94 & 85.01 & 84.75 & 91.17 & 91.40 & 91.51 & 81.18 & 80.29 & 79.03 & 85.76 & 85.57 & 85.10 \\
&\texttt{Dec-LoRA} (ER-$p_c=0.2$) & 84.89 & 84.73 & 83.98 & 91.63 & 91.40 & 90.94 & 80.28 & 79.31 & 77.64 & 85.60 & 85.15 & 84.19 \\
&\texttt{Dec-LoRA} (exponential graph) & 85.56 & 84.88 & 84.69 & 91.63 & 91.63 & 91.51 & 82.18 & 81.92 & 80.57 & 86.46 & 86.14 & 85.59 \\
\cline{2-14}
& \texttt{Dec-Adapters} (complete) & 88.38 & 88.07 & 86.78 & 93.23 & 92.66 & 92.55 & 84.10 & 83.32 & 82.63 & 88.57 & 88.01 & 87.32 \\
&\texttt{Dec-Adapters} (ring) & 83.60 & 79.52 & 61.43 & 92.20 & 92.09 & 91.40 & 75.62 & 64.44 & 57.73 & 83.81 & 78.68 & 70.19 \\
&\texttt{Dec-Adapters} (ER-$p_c=0.6$) & 86.93 & 85.78 & 85.06 & 93.12 & 92.32 & 91.63 & 81.59 & 79.86 & 78.35 & 87.21 & 85.99 & 85.01 \\
&\texttt{Dec-Adapters} (ER-$p_c=0.2$) & 85.19 & 82.22 & 70.60 & 92.43 & 91.86 & 91.28 & 78.57 & 74.64 & 70.25 & 85.40 & 82.91 & 79.71 \\
&\texttt{Dec-Adapters} (exponential graph) & 87.94 & 87.72 & 86.89 & 93.23 & 92.78 & 92.89 & 83.42 & 82.86 & 82.07 & 88.20 & 87.79 & 87.26 \\
\bottomrule
\end{tabular}
}
\end{center}
\end{table*}

\subsection{Ablation Study on LoRA Rank}
We conducted an ablation study on LoRA rank under the same setting in Table \ref{tab:com}, as shown in Table \ref{tab:rank}, and observed that higher ranks can improve performance of \texttt{Dec-LoRA}.
\begin{table*}[t]
\caption{Effect of different ranks for \texttt{Dec-LoRA}.}
\label{tab:rank}
\begin{center}
\resizebox{0.75\linewidth}{!}{
\begin{tabular}{lccccccccc}
\toprule 
 \multirow{2.5}{*}{Method (Rank)} &
 \multicolumn{2}{c}{QNLI} & \multicolumn{2}{c}{SST2} & \multicolumn{2}{c}{MNLI} &
 \multicolumn{2}{c}{QQP}&
 \multirow{2.5}{*}{Avg.}\\
 \cmidrule(lr){2-3} \cmidrule(lr){4-5} \cmidrule(lr){6-7} \cmidrule(lr){8-9} 
    &  Ring & ER & Ring & ER & Ring & ER & Ring & ER \\
\midrule
\textbf{\texttt{Dec-LoRA}} ($r=16$) 
& 90.13 & 90.04 & 92.89 & 93.46 & 85.13 & 84.98 & 87.59  & 87.40 & 88.95 \\
\textbf{\texttt{Dec-LoRA}} ($r=32$) 
& 90.48 & 90.23 & 94.27 & 93.00 & 85.24 & 85.31 & 87.63 & 87.10 & 89.16 \\
\textbf{\texttt{Dec-LoRA}} ($r=64$) 
& 90.19 & 89.91 & 93.58 & 93.46 & 85.46 & 85.13 & 87.67 & 87.58 & 89.12 \\
\bottomrule
\end{tabular}
}
\end{center}
\end{table*}

\subsection{Freezing Matrix $A$ in \texttt{Dec-LoRA}}
In this section, we adopt the idea of freezing matrix $A$ from~\cite{sun2024improving} and implement it in our decentralized setup, referring to it as \texttt{Dec-FFA-LoRA}.

We compare \texttt{Dec-LoRA} and \texttt{Dec-FFA-LoRA} in Table \ref{tab:FFA}, using $10$ clients, $5$ local updates, and non-i.i.d. data simulated via a Dirichlet distribution with $\alpha=0.5$. As shown, \texttt{Dec-LoRA} consistently achieves better performance.

While freezing $A$ may intuitively seem beneficial for mitigating discrepancy in federated learning, our results show it does not provide a clear advantage. This observation aligns with several prior works~\cite{yanfederated,singhal-etal-2025-fedex,koo-etal-2025-towards,singhal2025fedsb}, which also report that standard FL combined with LoRA tends to outperform \texttt{FFA-LoRA}.

For instance, \cite{yanfederated} notes:
``This is due to FFA-LoRA’s focus on differential privacy, as well as the discrepancy between our task setup and its original configuration, which employed manually partitioned data. Such partitioning does not fully reflect the complexity of real-world scenarios, leading to a mismatch in performance under our experimental settings.''

\begin{table*}[t]
\caption{Comparitive analysis of \texttt{Dec-LoRA} and \texttt{FFA-LoRA} in a decentralized setting.}
\label{tab:FFA}
\begin{center}
\resizebox{0.75\linewidth}{!}{
\begin{tabular}{lccccccccc}
\toprule 
 \multirow{2.5}{*}{Method (Rank)} &
 \multicolumn{2}{c}{QNLI} & \multicolumn{2}{c}{SST2} & \multicolumn{2}{c}{MNLI} &
 \multicolumn{2}{c}{QQP}&
 \multirow{2.5}{*}{Avg.}\\
 \cmidrule(lr){2-3} \cmidrule(lr){4-5} \cmidrule(lr){6-7} \cmidrule(lr){8-9} 
    &  Ring & ER & Ring & ER & Ring & ER & Ring & ER \\
\midrule
\textbf{\texttt{Dec-LoRA}}
& 91.23 & 91.63 & 94.61 & 94.27 & 85.94 & 85.60 & 85.10  & 86.76 & 89.21 \\
\textbf{\texttt{Dec-FFA-LoRA}} & 87.37 & 87.41 & 92.89 & 92.89 & 82.61 & 82.61 &  84.64 & 84.49 & 86.86\\
\midrule
\textbf{\texttt{Dec-LoRA}} & 90.55 & 88.32 & 93.81 & 93.58 & 85.02 & 83.55 & 86.40 & 84.62 & 88.23 \\
\textbf{\texttt{Dec-FFA-LoRA}} & 87.63 & 83.82 & 92.09 & 92.09 & 82.16  & 80.03 & 83.59 & 82.68 & 85.51\\
\bottomrule 
\end{tabular}
}
\end{center}
\end{table*}

\subsection{Mixing Matrix}\label{network_topology}
As previously discussed, in our decentralized framework, clients communicate exclusively along the edges of a fixed communication graph that connects $n$ nodes. Each edge in this graph is associated with a positive mixing weight. These weights are collectively represented by the mixing matrix $ Q \in \mathbb{R}^{n\times n} $. We assume that the mixing matrix $Q$ is symmetric and doubly stochastic, which is a common assumption in the literature to ensure the consensus~\cite{koloskova2020unified}. In this work, we utilize two widely used network topologies, which are described as follows:
\begin{itemize}
    \item \textbf{Ring topology} consists of nodes arranged in a closed-loop structure, where each node communicates only with its immediate neighbors, leading to a sparse mixing matrix \( Q \) with nonzero entries corresponding to these direct connections. While this structured and deterministic communication pattern simplifies theoretical analysis, the limited communication range can slow down information diffusion, potentially hindering the overall convergence speed of the learning process.  \textit{We use this challenging topology in many parts of our experiment section.}
    For the Ring topology, the mixing matrix is given by $Q = \frac{1}{3}(I + A)$,
    where $I$ is the identity matrix and $A$ is the adjacency matrix of the undirected ring. Moreover, $\beta$ (second largest magnitude among the eigenvalues of $Q$) can be written as \cite{xiao2004fast} $\beta \approx 1 - \mathcal{O}(\frac{1}{n^2})$.
    
    \item \textbf{Erd\H{o}s-R\'enyi (ER) topology} is a random graph model where each edge between nodes exists with an independent probability \( p_c \), but the connectivity structure remains fixed throughout training. The mixing matrix for the Erd\H{o}s-R\'enyi topology is defined as \( Q = I - \frac{2}{3 \lambda_{\max}(L)} L \), where \( L \) is the Laplacian matrix of an Erd\H{o}s-R\'enyi graph with edge probability \( p_c \). While a larger \( p_c \) results in a more connected network, facilitating faster information exchange, a smaller \( p_c \) leads to sparser connectivity, which may slow down convergence. \textit{This relationship has been tested for LLMs in the experiment section. }
    For ER topology, $\beta$ can be written as \cite{krivelevich2003largest} $\beta \approx 1 - \Theta(\frac{1}{\sqrt{n}})$, when \( p_c \gg \frac{\log n}{n} \).

\end{itemize}

\subsection{Impact of Number of Clients, Edge Probabilities, and Number of Local Updates}\label{app:impact}
In this section, we validate the empirical results presented in Section~\ref{comparitive2} using our theoretical findings. First, note that a smaller value of \( \beta \in (0,1) \) indicates a better-connected network and thus faster convergence. This is also reflected in our final upper bound, where \( M_A, M_B \propto \frac{\beta^2}{1 - \beta^2} \); a smaller \( \beta \) leads to a tighter bound.
\begin{itemize}
    \item \textbf{Number of clients ($n$):} For Ring and (ER) topologies, \( \beta \) can be approximated as \( 1 - \mathcal{O}(\frac{1}{n^2}) \) and \( 1 - \Theta(\frac{1}{\sqrt{n}}) \), respectively. As the number of clients \( n \) increases, \( \beta \) approaches 1, which slows down convergence. This effect is evident in Fig.~\ref{fig2} (a) and (d), where, for a fixed number of communication rounds, larger \( n \) leads to lower accuracies.

    \item \textbf{Edge probabilities ($p_c$):} As mentioned earlier, a larger \( p_c \) results in a more connected network, which corresponds to a lower \( \beta \) and thus faster convergence. This trend is illustrated in Fig.~\ref{fig2} (b) and (e), where a higher edge probability leads to improved accuracy for a fixed number of communication rounds.

    \item \textbf{Number of local updates ($K$):} Assuming that \( TK \), the total gradient complexity, is fixed, increasing \( K \) leads to a decrease in \( T \). Since the final convergence bound is proportional to \( \frac{1}{T^{1/3}} \), a smaller \( T \) results in slower convergence. This phenomenon is also evident in Fig.~\ref{fig2} (c) and (f), where a larger number of local updates corresponds to lower accuracies.

\end{itemize}

\subsection{Additional Details for Section \ref{bert}}\label{app.A}

\begin{table}[t]
\centering
\caption{Dataset descriptions and statistics.}\label{tab:glue_metric}
\small
\begin{tabular}{l|c|c|c}
\toprule
\textbf{Task} & \textbf{\# Train} & \textbf{\# Dev.} & \textbf{Metric}  \\ \midrule
MRPC      & 3,301 &  367   & F1 Score \\ 
SST-2     & 66,675 &  674  & Accuracy \\
QNLI      & 103,695 &   5,463 & Accuracy \\
QQP       & 360,210 &   40,430 & Accuracy \\ 
MNLI   & 388,774 &   9,815 & Accuracy \\
\bottomrule
\end{tabular}
\end{table}

For the experiments involving the BERT family, we utilize the Generalized Language Understanding Evaluation (GLUE) benchmark~\cite{wang2018glue}, which comprises various natural language understanding tasks. These include sentiment analysis (SST2~\cite{socher2013recursive}), similarity and paraphrasing tasks (MRPC, QQP~\cite{dagan2005pascal}), and natural language inference (MNLI, QNLI~\cite{williams2017broad, rajpurkar2018know}). The evaluation metrics for the GLUE benchmark are detailed in Table \ref{tab:glue_metric}. A learning rate of $1e-3$ and a batch size of $32$ are applied consistently across all tasks and methods. The results presented in Fig.~\ref{fig_bar} and Table~\ref{tab:quantization} are derived after $100$ communication rounds/epochs.

\subsection{Additional Details for Section \ref{large}}\label{app.B}
\textbf{Comparison with Other Methods.} For large-scale language models, we conduct experiments only on centralized LoRA and \texttt{Dec-LoRA}. Applying Adapters for fine-tuning large-scale models still requires a significant number of trainable parameters. For instance, applying Adapters to LLaMA-2-13B with a bottleneck size of $64$—the same as used for the BERT family—would require $50.33M$ trainable parameters, making it impractical for decentralized scenarios. As shown in Table \ref{tab:generation}, the number of trainable parameters remains relatively small when applying LoRA to LLaMA-2-7B. Additionally, since LLaMA-2-7B does not include bias terms, BitFit cannot be applied, as it updates only the bias parameters.

\begin{table*}[t]
\caption{A comparative analysis of centralized LoRA and \texttt{Dec-LoRA} with $10$ clients, using the LLaMA2-13B and OPT-2.7B models.}
\label{tab:generation2}
\begin{center}
\resizebox{\linewidth}{!}{
\begin{tabular}{ccccccccccccc}
\toprule 
\multirow{4}{*}{Rank} & \multicolumn{6}{c}{LLaMA-2-13B} & \multicolumn{6}{c}{OPT-2.7B}\\
\cmidrule(lr){2-7} \cmidrule(lr){8-13} 
   & \multicolumn{2}{c}{COPA} & \multicolumn{2}{c}{ReCoRD} &
 \multicolumn{2}{c}{SQuAD} &
 \multicolumn{2}{c}{SQuAD} &
 \multicolumn{2}{c}{BoolQ} &
 \multicolumn{2}{c}{ReCoRD}\\
 \cmidrule(lr){2-3} \cmidrule(lr){4-5} \cmidrule(lr){6-7} \cmidrule(lr){8-9} \cmidrule(lr){10-11} \cmidrule(lr){12-13} 
 & \texttt{LoRA} & \texttt{Dec-LoRA}$_{10}$  & \texttt{LoRA} & \texttt{Dec-LoRA}$_{10}$  & \texttt{LoRA} & \texttt{Dec-LoRA}$_{10}$  & \texttt{LoRA} & \texttt{Dec-LoRA}$_{10}$  & \texttt{LoRA} & \texttt{Dec-LoRA}$_{10}$  & \texttt{LoRA} & \texttt{Dec-LoRA}$_{10}$ \\
\midrule
$8$ & $92$ & $93$ & $84.2$ & $83.9$ & $92.24$ & $90.88$ &$81.93$ & $79.50$  &  $63.1$ & $63.6$ & $77.0$ & $75.8$  \\
\bottomrule
\end{tabular}
}
\end{center}
\end{table*}

For the experiments involving large-scale language models, we use a learning rate of $1e-4$ and a batch size of $2$ across all tasks and methods. All classification tasks within the SuperGLUE benchmark are restructured as language modeling tasks using the prompt-based fine-tuning approach outlined in~\cite{malladi2023fine}.The results shown in Table \ref{tab:generation} are obtained after completing $10$ communication rounds/epochs. The evaluation metrics are presented in Table \ref{tab:superglue_metric}.
\begin{table}[ht]
	\centering
	\caption{The utilized metrics for the SuperGLUE benchmark and generation tasks.}
	\label{tab:superglue_metric}
	\resizebox{0.2\textwidth}{!}{%
		\begin{tabular}{@{}cc@{}}
			\toprule
			Task Name & Metric                       \\ \midrule
                WIC         & F1\\
                BoolQ      & Accuracy\\
			COPA      & Accuracy                        \\
			ReCoRD      & F1                                    \\
			SQuAD    & F1     \\
                DROP     & F1\\
			\bottomrule
		\end{tabular}%
	}
\end{table}

Additional experiments conducted on LLaMA-2-13B and OPT-2.7B~\cite{zhang2022opt} are presented in Table \ref{tab:generation2} under the same setting.

\subsection{Learning Rate}
We explored different learning rates across many settings, and our observations showed that the method is not particularly sensitive to the choice of learning rate. In most cases, the best accuracy was consistently achieved with a learning rate of $1\text{e}{-3}$, so we adopted it for all experiments. We show the results for three learning rates in Table \ref{tab:learning-rate}.

\begin{table*}[t]
\caption{}
\label{tab:learning-rate}
\begin{center}
\resizebox{0.75\linewidth}{!}{
\begin{tabular}{lccccccccc}
\toprule 
 \multirow{2.5}{*}{Method (Learning Rate)} &
 \multicolumn{2}{c}{QNLI} & \multicolumn{2}{c}{SST2} & \multicolumn{2}{c}{MNLI} &
 \multicolumn{2}{c}{QQP}&
 \multirow{2.5}{*}{Avg.}\\
 \cmidrule(lr){2-3} \cmidrule(lr){4-5} \cmidrule(lr){6-7} \cmidrule(lr){8-9} 
    &  Ring & ER & Ring & ER & Ring & ER & Ring & ER \\
\midrule
\textbf{\texttt{Dec-LoRA}} ($lr=1e-3$) 
& 90.13 & 90.04 & 92.89 & 93.46 & 85.13 & 84.98 & 87.59  & 87.40 & 88.95 \\
\textbf{\texttt{Dec-LoRA}} ($lr=5e-4$) 
& 90.52 & 90.55 & 92.78  & 92.78 & 85.51 & 85.46 & 87.99 & 87.08 & 89.08 \\
\textbf{\texttt{Dec-LoRA}} ($lr=1e-4$) 
& 89.29 & 89.25 & 93.23 & 92.78 & 84.26 & 84.66 & 85.93 & 86.48 & 88.23 \\
\bottomrule
\end{tabular}
}
\end{center}
\end{table*}

\section{Convergence Analysis}\label{convergence}
\textbf{Notation.}
We employ the following notations of stacked matrices:
\begin{align}
&[A_i]:=\begin{bmatrix}\vspace{0.2cm}A_1\\\vspace{0.2cm} A_2\\\vspace{0.2cm}\vdots\\A_n\end{bmatrix}\in\mathbb{R}^{nr\times d_2}\quad\text{and}\quad
h_A^{(t)+k}:=\begin{bmatrix}\vspace{0.2cm}\nabla_A F_1(W_1^{(t)+k};\xi_1^{(t)+k})\\
\vspace{0.2cm}\nabla_A F_2(W_2^{(t)+k};\xi_2^{(t)+k})\\\vspace{0.2cm}\vdots\\ \nabla_A
F_n(W_n^{(t)+k};\xi_n^{(t)+k})\end{bmatrix}\in\mathbb{R}^{nr\times d_2}, \nonumber\\
& [B_i]:=\begin{bmatrix}\vspace{0.2cm}B_1 \\ \vspace{0.2cm} B_2\\ \vspace{0.2cm}\vdots\\B_n\end{bmatrix}\in\mathbb{R}^{nd_1\times r}\quad\text{and}\quad
h_B^{(t)+k}:=\begin{bmatrix}\vspace{0.2cm}\nabla_B F_1(W_1^{(t)+k};\xi_1^{(t)+k})\\
\vspace{0.2cm}\nabla_B F_2(W_2^{(t)+k};\xi_2^{(t)+k})\\\vspace{0.2cm}\vdots\\ \nabla_B
F_n(W_n^{(t)+k};\xi_n^{(t)+k})\end{bmatrix}\in\mathbb{R}^{nd_1\times r}. \nonumber
\end{align}
We define these quantities for simplicity in the proof.
\begin{align}
& \bar{A} := \frac{1}{n}\sum_{i=1}^n A_i, \quad \bar{B} := \frac{1}{n}\sum_{i=1}^n B_i, \quad
\hat{h}_A^{(t)} := \sum_{k=0}^{K-1}h_A^{(t)+k}, \quad\hat{h}_B^{(t)} := \sum_{k=0}^{K-1}h_B^{(t)+k},\nonumber\\
&g_A^{(t)+k}:=\frac{1}{n} \sum_{i=1}^n \nabla_A f_i(W_i^{(t)+k}),\quad \Tilde{g}_A^{(t)+k}:=\frac{1}{n} \sum_{i=1}^n \nabla_A F_i(W_i^{(t)+k};\xi_i^{(t)+k}),
\nonumber\\
& g_B^{(t)+k}:=\frac{1}{n} \sum_{i=1}^n \nabla_B f_i(W_i^{(t)+k}), \quad \Tilde{g}_B^{(t)+k}:=\frac{1}{n} \sum_{i=1}^n \nabla_B F_i(W_i^{(t)+k}; \xi_i^{(t)+k}),
\end{align}
where $A_i^{(t)+k}$, $B_i^{(t)+k}$, $W_i^{(t)+k}$ denote variables on node $c_i$ at local step $k$ and communication round $t$.
$\|A\|_F$ and $\langle A, B \rangle_F$ represent the Frobenius norm and the Frobenius inner product, respectively. $\|A\|_2 = \max_{x, \|x\|_2 = 1} \|Ax\|_2$ denotes the spectral norm of the matrix $A$. All matrix norms and inner products without subscripts are assumed to refer to the Frobenius norm and Frobenius inner product, respectively.

Before presenting the intermediate lemmas necessary for proving Theorem \ref{theorem}, we define the Kronecker product.  

\begin{definition}
For two matrices \( A \) of size \( d \times k \) and \( B \) of size \( p \times q \), their Kronecker product, denoted as \( A \otimes B \), is an \( (dp) \times (kq) \) matrix structured as:
$$
A \otimes B =
\begin{bmatrix}
a_{11}B & a_{12}B & \dots & a_{1n}B \\
a_{21}B & a_{22}B & \dots & a_{2n}B \\
\vdots  & \vdots  & \ddots & \vdots  \\
a_{m1}B & a_{m2}B & \dots & a_{mn}B
\end{bmatrix},
$$
\end{definition}
where $a_{ij}$ denotes the element located in the $i$-th row and $j$-th column of the matrix $A$.

\begin{lemma}\label{lemma:pre1}
For any matrices $ A, B \in \mathbb{R}^{d\times k}$ and $\alpha, \delta > 0$ we have
\begin{align}
    2\langle A, B\rangle &\leq \delta\|A\|^2+\delta^{-1}\|B\|^2,\nonumber\\
    \|A + B\|^2 &\leq (1 + \alpha)\|A\|^2 + (1 + \frac{1}{\alpha})\|B\|^2.\nonumber
\end{align}
\end{lemma}

\begin{lemma} For a set of arbitrary matrices $ A_1, \ldots, A_n $ such that $ A_i \in \mathbb{R}^{d\times k} $, we have
\begin{equation}
    \left\|\frac{1}{n}\sum_{i=1}^n A_i \right\|^2 \leq \frac{1}{n}\sum_{i=1}^n\|A_i\|^2.\nonumber
\end{equation}
\end{lemma}

\begin{lemma}\label{frobenius_inq}
    For matrices $A$ and $B$, the Frobenius norm satisfies the following inequality:
    \begin{align}
        \|AB\|_F \le \| A \|_2 \|B\|_F\nonumber.
    \end{align}
\end{lemma}

\begin{lemma}\label{beta}
\cite{yuan2016convergence} Assume that Assumption \ref{assm: mixing matrix} holds for the mixing matrix $Q$. Then, for any $N \geq 1$, we have:
\begin{align}
    \left\| Q^{N} - \frac{1}{n} 1_n 1_n^T \right\|_2 = \beta^N,
\end{align}
where $\beta$ is the second-largest absolute eigenvalue of $Q$.
\end{lemma}

\begin{lemma}\label{lemma:smooth}
    Under Assumption \ref{ass:smoothness}, each local objective function $f_i$ is $LC_B^2$-smooth with respect to $A$ when $B$ is fixed, and $LC_A^2$-smooth with respect to $B$ when $A$ is fixed.
\end{lemma}
\textit{Proof.}
Noticing that $\nabla_Af(BA) = B^T\nabla_Wf(BA)$, we have
\begin{align}
    \left\| \nabla_A f_i(BA) - \nabla_A f_i(BA^{\prime})\right\| &= \left\| B^T\nabla_W f_i(BA) - B^T\nabla_W f_i(BA^{\prime})\right\|\nonumber\\
    &
    =\left\|B\right\| \left\| \nabla_W f_i(BA) - \nabla_W f_i(BA^{\prime})\right\|\nonumber\\
    &
    \stackrel{(a)}{\le} LC_B  \left\|BA-BA^{\prime}\right\| \nonumber\\
    &
    \le LC_B^2  \left\|A-A^{\prime}\right\| .
\end{align}
In $(a)$ we used Assumptions \ref{ass:smoothness} and \ref{assm: bounded matrices}.
Similarly, we can prove that $f_i$ is $LC_A^2$-smooth with respect to $B$ when $A$ is fixed.
\hfill \ensuremath{\Box}

\begin{lemma}\label{lemma:general}
Under Assumptions \ref{ass:smoothness}, \ref{assm: bounded gradient}, \ref{assm: bounded matrices}, and for every $A$, $B$, $A'$, $B'$ we have
\begin{align}
    \mathbb{E} &\left\| \nabla_A f_i(BA) - \nabla_A f_i(B'A') \right\|^2 \le 4\mathbb{E} \left \|B-B' \right\|^2 \left(G^2+L^2C^2_BC_A^2\right) + 2\mathbb{E} \left \|A-A' \right\|^2 L^2 C_B^4,\nonumber\\
    \mathbb{E} &\left\| \nabla_A f_i(BA) - \nabla_A f_i(B'A') \right\|^2 \le 4\mathbb{E} \left \|A-A' \right\|^2 \left(G^2+L^2C^2_BC_A^2\right) + 2\mathbb{E} \left \|B-B' \right\|^2 L^2 C_A^4.
\end{align}
\end{lemma}
\textit{Proof.} We can start by writing that 
\begin{align}
 \mathbb{E} &\left\| \nabla_A f_i(BA) - \nabla_A f_i(B'A') \right\|^2  = \mathbb{E} \left\| B^T \nabla_W f_i(BA) - B'^T\nabla_W f_i(B'A') \right\|^2\nonumber\\
 & \stackrel{(a)}{\leq} 2\mathbb{E} \left\| B^T \nabla_W f_i(BA) - B'^T\nabla_W f_i(B'A) \right\|^2+2\mathbb{E} \left\| B'^T \nabla_W f_i(B'A) - B'^T\nabla_W f_i(B'A') \right\|^2\nonumber\\
 & \stackrel{(b)}{\leq} 4\mathbb{E} \left\| B^T \nabla_W f_i(BA) - B'^T\nabla_W f_i(BA) \right\|^2 +4\mathbb{E} \left\| B'^T \nabla_W f_i(B'A) - B'^T\nabla_W f_i(BA) \right\|^2\nonumber\\
 &\quad +2\mathbb{E} \left\| B'^T \nabla_W f_i(B'A) - B'^T\nabla_W f_i(B'A') \right\|^2\nonumber\\
 & \stackrel{(c)}{\leq} 4\mathbb{E} \left \|B-B' \right\|^2 \left(G^2+L^2C^2_BC_A^2\right) + 2\mathbb{E} \left \|A-A' \right\|^2 L^2 C_B^4,
\end{align}
where in $(a)$ and $(b)$, we apply the inequality $\|a+b\|^2 \leq 2\|a\|^2 + 2\|b\|^2$, in $(c)$, we utilize Assumptions \ref{ass:smoothness}, \ref{assm: bounded gradient}, and \ref{assm: bounded matrices}.
 Similarly, we can derive the bound for $\mathbb{E} \left\| \nabla_B f_i(BA) - \nabla_B f_i(B'A') \right\|^2$.
\hfill \ensuremath{\Box}

\begin{lemma}\label{lemma:5}
Under Assumptions \ref{assm: bounded gradient} and \ref{assm: bounded matrices} the averaged stacked vector of gradients is bounded, that is, 
\begin{align}
    \mathbb{E}\left\| \hat{h}_A^{(t)}\right\|^2 &\le
    n K^2C_B^2G^2, \nonumber\\\mathbb{E}\left\|\hat{h}_B^{(t)}\right\|^2 &\leq n K^2C_A^2G^2.
\end{align}
\end{lemma}
\textit{Proof.}
First, we notice that 
\begin{align}
    \nabla_A F_i(W_i^{(t)}; \xi_i^{(t)}) &= B_i^{(t)^T}\nabla_W F_i(W_i^{(t)}; \xi_i^{(t)}) \nonumber \\
     \nabla_B F_i(W_i^{(t)}; \xi_i^{(t)}) &=\nabla_W F_i(W_i^{(t)}; \xi_i^{(t)})A_i^{(t)^T}. \label{grad}
\end{align}
Recalling the definition of $\hat{h}_A^{(t)}$, we can write
\begin{align}
    \mathbb{E}\left\| \hat{h}_A^{(t)}\right\|^2 = \mathbb{E}\left\| \sum_{k=0}^{K-1}h_A^{(t)+k}\right\|^2 &\le K \sum_{k=0}^{K-1} \mathbb{E}\left\|h_A^{(t)+k} \right\|^2 \nonumber\\
    &= K \sum_{k=0}^{K-1}\sum\limits_{i=1}^{n}\mathbb{E}\left\|\nabla_A F_i(W_i^{(t)+k}; \xi_i^{(t)+k})\right\|^2 \nonumber\\
    &=K \sum_{k=0}^{K-1}\sum\limits_{i=1}^{n}\mathbb{E}\left\|\left(B_i^{(t)+k}\right)^T\nabla_W F_i(W_i^{(t)+k}; \xi_i^{(t)+k})\right\|^2 \nonumber\\
    &\stackrel{(a)}{\leq} KC_B^2 \sum_{k=0}^{K-1}\sum\limits_{i=1}^{n}\mathbb{E}\left\|\nabla_W F_i(W_i^{(t)+k}; \xi_i^{(t)+k})\right\|^2 \nonumber\\
    &\stackrel{(b)}{\leq} nK^2 C_B^2G^2 .
\end{align}
In $(a)$ and $(b)$ we used Assumptions \ref{assm: bounded matrices} and \ref{assm: bounded gradient}, respectively.
Similarly, we have:
$$
\mathbb{E}\left\|\hat{h}_B^{(t)}\right\|^2 \leq n K^2C_A^2G^2,
$$
which completes the proof.
\hfill \ensuremath{\Box}

\begin{lemma}\label{lemma:first_main}
Under Assumptions \ref{ass:smoothness}, \ref{assm: bounded gradient}, \ref{assm: bounded matrices}, \ref{assm: mixing matrix}, the total deviation from mean is vanishing for both $A$ and $B$, namely,   
\begin{align}
\frac{1}{n}\sum\limits_{i=1}^{n}\mathbb{E}\left\|A_i^{(t)}-\bar{A}^{(t)}\right\|^2 &\le  K^2\eta^2 M_A + a_0(t),\nonumber\\
\frac{1}{n}\sum\limits_{i=1}^{n}\mathbb{E} \left \|B_i^{(t)}-\bar{B}^{(t)}\right \|^2 &\leq K^2\eta^2M_B,
\end{align}
where $M_A=\frac{2(1+\beta^2)\beta^2 G^2 C_B^2}{(1 - \beta^2)^2}$, $M_B=\frac{2(1+\beta^2)\beta^2 G^2 C_A^2}{(1 - \beta^2)^2}$, $a_0(t) = \frac{1}{n}\rho^{t} \mathbb{E}\|[A_i^{(0)}] \|^2 $, and $\rho=\frac{1+\beta^2}{2}$.
\end{lemma}
\textit{Proof.}
From the definitions of $[A_i]$, $\hat{h}_A^{(t)}$, and \eqref{eq:updata}, \eqref{eq:average} we have
\begin{align}
   \left[A_i^{(t+1)}\right]=(Q\otimes I)\left(\left[A_i^{(t)}\right] - \eta\cdot \hat{h}_A^{(t)}\right), 
\end{align}
where 
$\otimes$ denotes the Kronecker product. Besides, letting $[\boldsymbol{\bar{A}}^{(t)}]=[\bar{A}^{(t)} ; \cdots ; \bar{A}^{(t)}] \in \mathbb{R}^{n r \times k}$, it follows that 
\begin{align}
\left[\bar{A}^{(t+1)}\right]=\frac{1}{n}\left(1_n 1_n^T \otimes I\right)\left[A_i^{(t+1)}\right]=\frac{1}{n}\left(1_n 1_n^T \otimes I\right)\left(\left[A_i^{(t)}\right]-\eta \hat{h}_A^{(t)}\right) .
\end{align}

We now derive a recursion for $\delta_A^{(t+1)}$: 
\begin{align}
\delta_A^{(t+1)} & =\left[A_i^{(t+1)}\right]-\left[\bar{A}^{(t+1)}\right] \nonumber \\
& =(Q \otimes I)\left(\left[A_i^{(t)}\right]-\eta \hat{h}_A^{(t)}\right)-\frac{1}{n}\left(1_n 1_n^T \otimes I\right)\left(\left[A_i^{(t)}\right]-\eta \hat{h}_A^{(t)}\right) \nonumber\\
& =\left(\left(Q-\frac{1}{n} 1_n 1_n^T\right) \otimes I\right)\left(\left[A_i^{(t)}\right]-\eta \hat{h}_A^{(t)}\right) .
\end{align}

We define $P:=\left(Q-\frac{1}{n} 1_n 1_n^T\right) \otimes I$. Since $P$ projects out the consensus component, we have:
\begin{align}
    &P\left[\bar{A}^{(t)}\right]=0 \quad \Rightarrow \quad P\left[A_i^{(t)}\right]=P\left(\delta_A^{(t)}+\bar{A}^{(t)}\right)=P\left(\delta_A^{(t)}\right),
\end{align}
which results in
\begin{align}
    \delta_A^{(t+1)}=P \delta_A^{(t)}-\eta P \hat{h}_A^{(t)}.
\end{align}
Taking expectations and applying Young's inequality, we have:
\begin{align}
\mathbb{E}\left\|\delta_A^{(t+1)}\right\|^2 \leq(1+\alpha) \beta^2 \mathbb{E}\left\|\delta_A^{(t)}\right\|^2+(1+1 / \alpha) \eta^2 \beta^2 \mathbb{E}\left\|\hat{h}_A^{(t)}\right\|^2.
\end{align}
Choosing $\alpha=\frac{1-\beta^2}{2 \beta^2}$ gives:
\begin{align}
    \rho:=(1+\alpha) \beta^2=\frac{1+\beta^2}{2}<1.
\end{align}
Unrolling the recursion and using Lemma \ref{lemma:5}, we have:
\begin{align}
  \frac{1}{n} \sum_{i=1}^n \mathbb{E}\left\|A_i^{(t)}-\bar{A}^{(t)}\right\|^2 \leq K^2 \eta^2\underbrace{\frac{2\left(1+\beta^2\right) \beta^2G^2 C_B^2}{\left(1-\beta^2\right)^2}}_{:=M_A}+\underbrace{\frac{1}{n} \rho^t \mathbb{E}\left\|\left[A_i^{(0)}\right]\right\|^2}_{:=a_0(t)}=K^2  \eta^2 M_A+a_0(t).
\end{align}
Similarly, we have:
\begin{align}
\frac{1}{n}\sum\limits_{i=1}^{n}\mathbb{E} \left \|B_i^{(t)}-\bar{B}^{(t)}\right \|^2 \leq K^2\eta^2 \underbrace{\frac{2\left(1+\beta^2\right) \beta^2G^2 C_A^2}{\left(1-\beta^2\right)^2}}_{:=M_B}+\frac{1}{n}\rho^t \mathbb{E}\underbrace{\left\|\left[B_i^{(0)}\right] \right\|^2}_{=0} = K^2  \eta^2 M_B,
\end{align}
which completes the proof.
\hfill \ensuremath{\Box}

\begin{lemma}\label{lemma:bounds}
Under Assumptions \ref{ass:smoothness}, \ref{assm: bounded gradient}, \ref{assm: bounded matrices}, \ref{assm: mixing matrix}, we have the following bounds
\begin{align}
    \mathbb{E}\left\|g_
    A^{(t)+k}-\nabla_Af(\bar{B}^{(t)}\bar{A}^{(t)})\right\|^2 \le K^2\eta^2\Tilde{M}_A+K^2\eta^2\hat{M}_A+J_A a_0(t)\nonumber,\\
    \mathbb{E}\left\|g_
    B^{(t)+k}-\nabla_Bf(\bar{B}^{(t)}\bar{A}^{(t)})\right\|^2 \le K^2\eta^2\Tilde{M}_B+K^2\eta^2\hat{M}_B+J_B a_0(t),
\end{align}

where $\Tilde{M}_A = 8M_B\left(G^2+L^2C^2_BC_A^2\right) + 4M_A L^2 C_B^4$, $\Tilde{M}_B=8M_A\left(G^2+L^2C^2_BC_A^2\right) + 4M_B L^2 C_A^4$, $J_A = 4L^2C_B^4$, $J_B= 8(G^2+4C_A^2C_B^2L^2)$, $\hat{M}_A = 8 G^2 C_A^2 \left(G^2+L^2C^2_BC_A^2\right) + 4 G^2 C_B^6$, and $\hat{M}_B = 8 G^2 C_B^2 \left(G^2+L^2C^2_BC_A^2\right) + 4 G^2 C_A^6$.
\end{lemma}
\textit{Proof.} We begin by writing that
\begin{align}
    \mathbb{E}\left\|g_
    A^{(t)+k}-\nabla_Af(\bar{B}^{(t)}\bar{A}^{(t)})\right\|^2 \le 2\mathbb{E}\left\|g_
    A^{(t)+k}-g_
    A^{(t)}\right\|^2 + 2\mathbb{E}\left\|g_
    A^{(t)}-\nabla_Af(\bar{B}^{(t)}\bar{A}^{(t)})\right\|^2.\label{lemma10:eq1}
\end{align}
Now, we derive upper bounds for the two terms in the above. Recalling the definition of $g_A^{(t)}$, we can write
\begin{align}
    2\mathbb{E}&\left\|g_
    A^{(t)}-\nabla_Af(\bar{B}^{(t)}\bar{A}^{(t)})\right\|^2 = 2\mathbb{E}\left\|\frac{1}{n}\sum_{i=1}^n  \left(\nabla_A f_i(B_{i}^{(t)} A_{i}^{(t)}) - \nabla_A f_i(\bar{B}^{(t)}\bar{A}^{(t)})\right)\right\|^2 \nonumber\\
    & \le \frac{2}{n}\sum_{i=1}^n \mathbb{E}\left\| \nabla_A f_i(B_{i}^{(t)} A_{i}^{(t)}) - \nabla_A f_i(\bar{B}^{(t)}\bar{A}^{(t)})\right\|^2\nonumber\\
    & \stackrel{(a)}{\leq} \frac{8}{n}\sum_{i=1}^n \mathbb{E} \left\|B_{i}^{(t)}-\bar{B}^{(t)} \right\|^2\left(G^2+L^2C^2_BC_A^2\right) + \frac{4}{n}\sum_{i=1}^n\mathbb{E} \left\|A_{i}^{(t)}-\bar{A}^{(t)} \right\|^2 L^2 C_B^4\nonumber\\
    & \stackrel{(b)}{\leq} 8M_B\left(G^2+L^2C^2_BC_A^2\right)K^2\eta^2 + 4L^2 C_B^4\left(K^2 \eta^2M_A+a_0(t)\right)\nonumber\\ 
    &=K^2\eta^2 \underbrace{\left(8M_B\left(G^2+L^2C^2_BC_A^2\right) + 4M_A L^2 C_B^4\right)}_{:=\Tilde{M}_A}
    +\underbrace{4L^2C_B^4}_{:=J_A} a_0(t) = K^2\eta^2\Tilde{M}_A+J_A a_0(t),\label{lemma10:eq2}
\end{align}
where in $(a)$, we utilize Lemma \ref{lemma:general}, and in $(b)$, we invoke Lemma \ref{lemma:first_main}. Next, we write
\begin{align}
    2\mathbb{E}&\left\|g_
    A^{(t)+k}-g_A^{(t)}\right\|^2=2\mathbb{E}\left\|\frac{1}{n}\sum_{i=1}^n  \left(\nabla_A f_i(B_{i}^{(t)+k} A_{i}^{(t)+k}) - \nabla_A f_i(B_{i}^{(t)} A_{i}^{(t)})\right)\right\|^2 \nonumber\\
    & \stackrel{(a)}{\leq} \frac{8}{n}\sum_{i=1}^n \mathbb{E} \left\|B_{i}^{(t)+k}-B_{i}^{(t)} \right\|^2\left(G^2+L^2C^2_BC_A^2\right) + \frac{4}{n}\sum_{i=1}^n\mathbb{E} \left\|A_{i}^{(t)+k}-A_{i}^{(t)} \right\|^2 L^2 C_B^4\nonumber\\
    &\stackrel{(b)}{\leq}  K^2\eta^2\underbrace{\left(8 G^2 C_A^2 \left(G^2+L^2C^2_BC_A^2\right) + 4 G^2 C_B^6\right)}_{:=\hat{M}_A} =K^2\eta^2\hat{M}_A\label{lemma10:eq3},
\end{align}
where we use Lemma \ref{lemma:general} and Lemma \ref{lemma:5} in $(a)$ and $(b)$, respectively. Combining \ref{lemma10:eq2} and \ref{lemma10:eq3} with \ref{lemma10:eq1}, we have
\begin{align}
    \mathbb{E}\left\|g_
    A^{(t)+k}-\nabla_Af(\bar{B}^{(t)}\bar{A}^{(t)})\right\|^2 \le K^2\eta^2\Tilde{M}_A+K^2\eta^2\hat{M}_A+J_A a_0(t).
\end{align}
Similarly, we can obtain a bound on $\mathbb{E}\|g_
    B^{(t)+k}-\nabla_Bf(\bar{B}^{(t)}\bar{A}^{(t)})\|^2$ in terms of $\Tilde{M}_B$, $J_B$, and $\hat{M}_B$.\hfill \ensuremath{\Box}

\begin{theorem}
Let Assumptions \ref{ass:smoothness}, \ref{assm: bounded gradient}, \ref{assm: bounded matrices}, and \ref{assm: mixing matrix} hold. Suppose in \eqref{eq:updata} we set $\eta=\frac{1}{KT^{\frac{1}{2}}}$. Then, after $T$ communication rounds, the following holds  
\begin{align}
    \frac{1}{T}\sum_{t=0}^{T-1} &\left(\mathbb{E}\left\|\nabla_A f(\bar{B}^{(t)}\bar{A}^{(t)})\right\|^2+\mathbb{E}\left\|\nabla_B f(\bar{B}^{(t)}\bar{A}^{(t)})\right\|^2\right) \le \mathcal{O}\left(\tfrac{\Delta_f+\Tilde{M}_A+\Tilde{M}_B}{T^{\frac{1}{2}}}+\tfrac{M+\hat{M}_A+\hat{M}_B}{T}\right.\nonumber\\
    &\hspace{5cm}+\left.\tfrac{(J_A+J_B)\mathbb{E}\|[A_i^{(0)}] \|^2}{T(1-\beta^2)}+\tfrac{L(C_A^4+C_B^4)G^2}{T^{\frac{1}{2}}}\right).
\end{align}

where $ \Delta_f = \mathbb{E}f(\bar{B}^{(0)}\bar{A}^{(0)})-\mathbb{E}f(\bar{B}^{(T)}\bar{A}^{(T)}) $, $M = 2C_A^4 C_B^2 L^2 G^2 + 2C_A^2 G^4 $, and $\Tilde{M}_A$, $\Tilde{M}_B$, $\hat{M}_A$, $\hat{M}_B$, $J_A$, $J_B$ are defined in Lemma \ref{lemma:bounds}.

\end{theorem}
\textit{Proof.}
Using smoothness for $A$ from Lemma \ref{lemma:smooth}, and noticing that 
\begin{align}
    \bar{A}^{(t)} = \frac{1}{n}\sum_{i,j=1}^n q_{ij} \left(A_j^{(t-1)}- \eta\sum_{k=0}^{K-1}\nabla_A F_i(W_i^{(t-1)+k};\xi_i^{(t-1)+k})\right) = \bar{A}^{(t-1)} - \eta\sum_{k=0}^{K-1} \Tilde{g}_A^{(t-1)+k},
\end{align}
we can write
\begin{align}
    \mathbb{E}&f(\bar{B}^{(t)}\bar{A}^{(t)})\le \mathbb{E}f(\bar{B}^{(t)}\bar{A}^{(t-1)})+\mathbb{E}\left\langle \nabla_A f(\bar{B}^{(t)}\bar{A}^{(t-1)}), \bar{A}^{(t)}-\bar{A}^{(t-1)}\right\rangle + \frac{LC_B^2}{2}\mathbb{E}\left\|\bar{A}^{(t)}-\bar{A}^{(t-1)}\right\|^2\nonumber\\
    &\stackrel{(a)}{=} \mathbb{E}f(\bar{B}^{(t)}\bar{A}^{(t-1)})+\mathbb{E}\left\langle \nabla_A f(\bar{B}^{(t)}\bar{A}^{(t-1)}), -\eta\sum_{k=0}^{K-1} g_A^{(t-1)+k}\right\rangle + \frac{LC_B^2\eta^2}{2}\mathbb{E}\left\|\sum_{k=0}^{K-1} \Tilde{g}_A^{(t-1)+k}\right\|^2\nonumber\\
    &\stackrel{(b)}{\le} 
    \mathbb{E}f(\bar{B}^{(t)}\bar{A}^{(t-1)}) +\frac{LC_B^4K^2G^2\eta^2}{2}\nonumber\\
    &
    \quad -\eta \mathbb{E}\left\langle \underbrace{\nabla_A f(\bar{B}^{(t)}\bar{A}^{(t-1)}) -\nabla_A f(\bar{B}^{(t-1)}\bar{A}^{(t-1)})}_{:=\Delta_A}+\nabla_A f(\bar{B}^{(t-1)}\bar{A}^{(t-1)}), \sum_{k=0}^{K-1} g_A^{(t-1)+k}\right\rangle \nonumber\\
    &=\mathbb{E}f(\bar{B}^{(t)}\bar{A}^{(t-1)}) -\eta \sum_{k=0}^{K-1}\mathbb{E}\left\langle \Delta_A,  g_A^{(t-1)+k}\right\rangle -\eta \sum_{k=0}^{K-1}\mathbb{E}\left\langle \nabla_A f(\bar{B}^{(t-1)}\bar{A}^{(t-1)}), g_A^{(t-1)+k}\right\rangle\nonumber\\
    &\quad+\frac{LC_B^4K^2G^2\eta^2}{2}\nonumber\\
    &
    \stackrel{(c)}{\le} 
    \mathbb{E}f(\bar{B}^{(t)}\bar{A}^{(t-1)}) +\eta \sum_{k=0}^{K-1}\left( 2\mathbb{E} \left\| \Delta_A\right\|^2 + \tfrac{1}{8}\mathbb{E} \left\|g_A^{(t-1)+k} \right\|^2\right)+\frac{LC_B^4K^2G^2\eta^2}{2}\nonumber\\
    &
     -\eta\sum_{k=0}^{K-1} \mathbb{E}\left\langle \nabla_A f(\bar{B}^{(t-1)}\bar{A}^{(t-1)}),  g_A^{(t-1)+k}-\nabla_A f(\bar{B}^{(t-1)}\bar{A}^{(t-1)})+\nabla_A f(\bar{B}^{(t-1)}\bar{A}^{(t-1)})\right\rangle\nonumber\\
     &
    \stackrel{(d)}{\le} 
    \mathbb{E}f(\bar{B}^{(t)}\bar{A}^{(t-1)})+\frac{LC_B^4K^2G^2\eta^2}{2}\nonumber \\
    &\quad+\eta \sum_{k=0}^{K-1}\left( 2\mathbb{E} \left\| \Delta_A\right\|^2 +\tfrac{1}{4}\mathbb{E}\left\|\nabla_A f(\bar{B}^{(t-1)}\bar{A}^{(t-1)})\right\|^2 + \tfrac{1}{4}\mathbb{E}\left\|g_A^{(t-1)+k}-\nabla_A f(\bar{B}^{(t-1)}\bar{A}^{(t-1)})\right\|^2\right)\nonumber\\
    &\quad +\eta\sum_{k=0}^{K-1}\left(- \tfrac{3}{4}\mathbb{E}\left\|\nabla_A f(\bar{B}^{(t-1)}\bar{A}^{(t-1)})\right\|^2 + \mathbb{E}\left\|g_A^{(t-1)+k}-\nabla_A f(\bar{B}^{(t-1)}\bar{A}^{(t-1)})\right\|^2 \right)\nonumber\\
    &
    = 
    \mathbb{E}f(\bar{B}^{(t)}\bar{A}^{(t-1)}) +2K\eta \mathbb{E}\left\|\Delta_A\right\|^2-\tfrac{\eta K}{2}\mathbb{E}\left\|\nabla_A f(\bar{B}^{(t-1)}\bar{A}^{(t-1)})\right\|^2+\frac{LC_B^4K^2G^2\eta^2}{2} \nonumber\\&
    \quad  + \tfrac{5\eta}{4}\sum_{k=0}^{K-1}\mathbb{E}\left\|g_A^{(t-1)+k}-\nabla_A f(\bar{B}^{(t-1)}\bar{A}^{(t-1)})\right\|^2,\label{eq:6}
\end{align}

where in $(a)$ we used $\mathbb{E}[\Tilde{g}_A^{(t-1)}] = g_A^{(t-1)}$, in $(b)$ we applied Lemma \ref{lemma:5}, in $(c)$ we employed the inequality $\langle a, b\rangle \leq \frac{1}{8}\|a\|^2 + 2\|b\|^2$, and in $(d)$ we utilized the inequalities $\langle a, b\rangle \leq \frac{1}{4}\|a\|^2 + \|b\|^2$ and $\|a+b\|^2 \leq 2\|a\|^2 + 2\|b\|^2$.

Now, we bound the following term in the above inequality
\begin{align}
    \mathbb{E}\left\|\Delta_A\right\|^2 &= \mathbb{E}\left\|\nabla_A f(\bar{B}^{(t)}\bar{A}^{(t-1)}) -\nabla_A f(\bar{B}^{(t-1)}\bar{A}^{(t-1)})\right\|^2 \nonumber\\
    &= \mathbb{E}\left\|\bar{B}^{(t)^T}\nabla_W f(\bar{B}^{(t)}\bar{A}^{(t-1)}) -\bar{B}^{(t-1)^T}\nabla_W f(\bar{B}^{(t-1)}\bar{A}^{(t-1)})\right\|^2\nonumber\\
    &\le2\mathbb{E}\left\|\bar{B}^{(t)^T}\nabla_W f(\bar{B}^{(t)}\bar{A}^{(t-1)}) -\bar{B}^{(t)^T}\nabla_W f(\bar{B}^{(t-1)}\bar{A}^{(t-1)})\right\|^2\nonumber\\
    &
    \quad+2\mathbb{E}\left\|\bar{B}^{(t)^T}\nabla_W f(\bar{B}^{(t-1)}\bar{A}^{(t-1)}) -\bar{B}^{(t-1)^T}\nabla_W f(\bar{B}^{(t-1)}\bar{A}^{(t-1)})\right\|^2\nonumber\\
    & \le K^2\eta^2\underbrace{(2C_A^4 C_B^2 L^2 G^2 + 2C_A^2 G^4)}_{:=M} =K^2 \eta^2 M .\label{eq:7}
\end{align}
Using Lemma \ref{lemma:bounds} and \eqref{eq:7} in \eqref{eq:6}, we obtain
\begin{align}\label{eq:8}
    \mathbb{E}f(\bar{B}^{(t)}\bar{A}^{(t)})
    &\le \mathbb{E}f(\bar{B}^{(t)}\bar{A}^{(t-1)}) +2\eta^3 K^3 M-\tfrac{\eta K}{2}\mathbb{E}\left\|\nabla_A f(\bar{B}^{(t-1)}\bar{A}^{(t-1)})\right\|^2+\frac{LC_B^4K^2G^2\eta^2}{2} \nonumber\\&
    \quad  + \tfrac{5\eta K}{4}\left(K^2\eta^2\Tilde{M}_A+K^2\eta^2\hat{M}_A+J_A a_0(t-1)\right).
\end{align}
Using smoothness for $B$ from Lemma \ref{lemma:smooth}, we can write
\begin{align}
    \mathbb{E}&f(\bar{B}^{(t)}\bar{A}^{(t-1)})\le \mathbb{E}f(\bar{B}^{(t-1)}\bar{A}^{(t-1)})+\mathbb{E}\left\langle \nabla_B f(\bar{B}^{(t-1)}\bar{A}^{(t-1)}), \bar{B}^{(t)}-\bar{B}^{(t-1)}\right\rangle \nonumber\\
    & \quad+ \frac{LC_A^2}{2}\mathbb{E}\left\|\bar{B}^{(t)}-\bar{B}^{(t-1)}\right\|^2\nonumber\\
    &= \mathbb{E}f(\bar{B}^{(t-1)}\bar{A}^{(t-1)})+\mathbb{E}\left\langle \nabla_B f(\bar{B}^{(t-1)}\bar{A}^{(t-1)}), -\eta\sum_{k=0}^{K-1}  g_B^{(t-1)+k}\right\rangle+ \tfrac{LC_A^2\eta^2}{2}\mathbb{E}\left\|\sum_{k=0}^{K-1}\Tilde{g}_B^{(t-1)+k}\right\|^2\nonumber\\
    &
    \le 
    \mathbb{E}f(\bar{B}^{(t-1)}\bar{A}^{(t-1)})+ \frac{LC_A^4 K^2G^2\eta^2}{2}
     \nonumber\\
    &
    \quad -\eta \sum_{k=0}^{K-1}\mathbb{E}\left\langle \nabla_B f(\bar{B}^{(t-1)}\bar{A}^{(t-1)}),  g_B^{(t-1)+k} -\nabla_B f(\bar{B}^{(t-1)}\bar{A}^{(t-1)})+\nabla_B f(\bar{B}^{(t-1)}\bar{A}^{(t-1)})\right\rangle \nonumber\\
    &
    \le\mathbb{E}f(\bar{B}^{(t-1)}\bar{A}^{(t-1)}) + \frac{LC_A^4 K^2G^2\eta^2}{2}\nonumber\\
    &
    \quad+ \frac{\eta}{2}\sum_{k=0}^{K-1}\left(-\mathbb{E}\left\|\nabla_B f(\bar{B}^{(t-1)}\bar{A}^{(t-1)})\right\|^2 +\mathbb{E} \left\| g_B^{(t-1)+k} -\nabla_B f(\bar{B}^{(t-1)}\bar{A}^{(t-1)})\right\|^2\right)\nonumber\\
    &\le\mathbb{E}f(\bar{B}^{(t-1)}\bar{A}^{(t-1)}) - \tfrac{\eta K}{2}\mathbb{E}\left\|\nabla_B f(\bar{B}^{(t-1)}\bar{A}^{(t-1)})\right\|^2+\frac{LC_A^4 K^2G^2\eta^2}{2}\nonumber\\
    & \quad + \tfrac{K\eta}{2} \left( K^2\eta^2\Tilde{M}_B+K^2\eta^2\hat{M}_B+J_B a_0(t-1) \right).
    \label{eq:9}
\end{align}
Summing up \eqref{eq:8} and \eqref{eq:9}, we obtain
\begin{align}
    \mathbb{E}&f(\bar{B}^{(t)}\bar{A}^{(t)})
    \le \mathbb{E}f(\bar{B}^{(t-1)}\bar{A}^{(t-1)}) - \tfrac{\eta K}{2} \left(\mathbb{E}\left\|\nabla_A f(\bar{B}^{(t-1)}\bar{A}^{(t-1)})\right\|^2+\mathbb{E}\left\|\nabla_B f(\bar{B}^{(t-1)}\bar{A}^{(t-1)})\right\|^2\right) \nonumber\\
    &
    \hspace{1.5cm}+ \left(\tfrac{5\Tilde{M}_A+2\Tilde{M}_B}{4}\right)K^3\eta^3+\left(\tfrac{5\hat{M}_A+2\hat{M}_B}{4}+2M\right)K^3\eta^3+\left(\tfrac{5 J_A+2J_B}{4}\right)K\eta a_0(t-1)\nonumber\\
    &\hspace{1.8cm}+\left(\tfrac{C_A^4+C_B^4}{2}\right)LG^2K^2\eta^2.
\end{align}
Then, it follows that 
\begin{align}
    \mathbb{E}&\left\|\nabla_A f(\bar{B}^{(t-1)}\bar{A}^{(t-1)})\right\|^2+\mathbb{E}\left\|\nabla_B f(\bar{B}^{(t-1)}\bar{A}^{(t-1)})\right\|^2
    \le \tfrac{2\left(\mathbb{E}f(\bar{B}^{(t-1)}\bar{A}^{(t-1)})-\mathbb{E}f(\bar{B}^{(t)}\bar{A}^{(t)})\right)}{K\eta}\nonumber\\
    &\hspace{0.8cm}+ \left(\tfrac{5\Tilde{M}_A+2\Tilde{M}_B}{2}\right)K^2\eta^2+\left(\tfrac{5\hat{M}_A+2\hat{M}_B}{2}+4M\right)K^2\eta^2+\left(\tfrac{5 J_A+2J_B}{2}\right) a_0(t-1)\nonumber\\
    &\hspace{1.1cm}+\left(C_A^4+C_B^4\right)LG^2K\eta.\label{eq:10}
\end{align}
By repeatedly applying \eqref{eq:10} for different values of $t$ and summing the results, we obtain
\begin{align}
     \frac{1}{T}\sum_{t=0}^{T-1}&\left(\mathbb{E}\left\|\nabla_A f(\bar{B}^{(t)}\bar{A}^{(t)})\right\|^2+\mathbb{E}\left\|\nabla_B f(\bar{B}^{(t)}\bar{A}^{(t)})\right\|^2\right)
    \le \tfrac{2\left(\mathbb{E}f(\bar{B}^{(0)}\bar{A}^{(0)})-\mathbb{E}f(\bar{B}^{(T)}\bar{A}^{(T)})\right)}{TK\eta}\nonumber\\
    &\quad+ \left(\tfrac{5\Tilde{M}_A+2\Tilde{M}_B}{2}\right)K^2\eta^2+\left(\tfrac{5\hat{M}_A+2\hat{M}_B}{2}+4M\right)K^2\eta^2+\left(\tfrac{5 J_A+2J_B}{2}\right) \tfrac{\sum_{t=0}^{T-1}a_0(t)}{T}\nonumber\\
    &\quad\quad+\left(C_A^4+C_B^4\right)LG^2K\eta\nonumber\\
    &\hspace{-0.5cm}\stackrel{(a)}{\le} \frac{2\Delta_f}{TK\eta}
     + \left(\tfrac{5\Tilde{M}_A+2\Tilde{M}_B}{2}\right)K^2\eta^2+\left(\tfrac{5\hat{M}_A+2\hat{M}_B}{2}+4M\right)K^2\eta^2+\left(\tfrac{5 J_A+2J_B}{2}\right) \tfrac{\mathbb{E}\|[A_i^{(0)}] \|^2}{nT(1-\rho)}\nonumber\\
     &\quad+\left(C_A^4+C_B^4\right)LG^2K\eta.
\end{align}

where in $(a)$, we use that $ \sum_t a_0(t) \le \frac{\mathbb{E} \|[A_i^{(0)}]\|^2}{n(1-\rho)} $. Now, by setting $\eta = \frac{1}{KT^{\frac{1}{2}}}$, we obtain
\begin{align}
    \frac{1}{T}\sum_{t=0}^{T-1} &\left(\mathbb{E}\left\|\nabla_A f(\bar{B}^{(t)}\bar{A}^{(t)})\right\|^2+\mathbb{E}\left\|\nabla_B f(\bar{B}^{(t)}\bar{A}^{(t)})\right\|^2\right) \le \tfrac{2\Delta_f+2.5\Tilde{M}_A+\Tilde{M}_B}{T^{\frac{1}{2}}}+\tfrac{4M+2.5\hat{M}_A+\hat{M}_B}{T}\nonumber\\
    &\hspace{5cm}+\tfrac{(2.5J_A+J_B)\mathbb{E}\|[A_i^{(0)}] \|^2}{nT(1-\rho)}+\tfrac{L(C_A^4+C_B^4)G^2}{T^{\frac{1}{2}}}.
\end{align}
\hfill \ensuremath{\Box}

\begin{remark}
    In the following, we specify the source of each error term in the final bound in Theorem \ref{theorem}.
    \begin{align}
     \frac{1}{T}\sum_{t=0}^{T-1}&\left(\mathbb{E}\left\|\nabla_A f(\bar{B}^{(t)}\bar{A}^{(t)})\right\|^2+\mathbb{E}\left\|\nabla_B f(\bar{B}^{(t)}\bar{A}^{(t)})\right\|^2\right)
    \le \underbrace{\frac{2\Delta_f}{TK\eta} +\left(C_A^4+C_B^4\right)LG^2K\eta}_{\text {Error with full synchronization}}\nonumber\\
    &+ \underbrace{\left(\tfrac{5\tilde{M}_A+2\tilde{M}_B}{2}\right)K^2\eta^2+\left(\tfrac{5\hat{M}_A+2\hat{M}_B}{2}\right)K^2\eta^2}_{\text{Error due to consensus and local updates}}+\underbrace{4MK^2\eta^2}_{\text{Error due to bilinear structure of LoRA}}\nonumber\\&+\underbrace{\left(\tfrac{5 J_A+2J_B}{2}\right) \tfrac{\sum_{t=0}^{T-1}a_0(t)}{T}}_{\text{Error due to Matrix $A$ initialization}}.
\end{align}
\end{remark}

\section{Limitations}\label{limitation}
As shown in Section~\ref{hetero}, the \texttt{Dec-LoRA} algorithm can experience performance degradation under data heterogeneity. This issue tends to become more pronounced as the number of clients and local updates increases. In the context of federated LLMs, methods such as~\cite{babakniya2023slora,yan2024federa} attempt to mitigate this challenge. Similarly, research like~\cite{liu2024decentralized,ghiasvand2025robust} aims to address data heterogeneity in decentralized learning settings more generally. Investigating these existing approaches or developing new algorithms to tackle this issue remains a promising avenue for future research.


\end{document}